\documentclass[preprint,12pt]{elsarticle}

\usepackage{amssymb}
\usepackage{amsmath} 
\usepackage{subfigure}
\usepackage{multirow}
\usepackage{float}
\usepackage{url} 

\journal{Pattern Recognition}
\begin{document}

\begin{frontmatter}



\title{Incremental Top-$k$ List Comparison Approach to Robust Multi-Structure Model Fitting}


\author{Hoi Sim Wong, Tat-Jun Chin, Jin Yu and David Suter}

\address{School of Computer Science, The University of Adelaide, South Australia, 5005}

\begin{abstract}
Random hypothesis sampling lies at the core of many popular robust fitting techniques such as RANSAC. In this paper, we propose a novel hypothesis 
sampling scheme based on incremental computation of distances between partial rankings (top-$k$ lists) derived from residual sorting information. Our 
method simultaneously (1) guides the sampling such that hypotheses corresponding to all true structures can be quickly retrieved and (2) filters the 
hypotheses such that only a small but very promising subset remain. This permits the usage of simple agglomerative clustering on the surviving 
hypotheses for accurate model selection. The outcome is a highly efficient multi-structure robust estimation technique. Experiments on synthetic and 
real data show the superior performance of our approach over previous methods.
\end{abstract}

\begin{keyword}
Model fitting, guided sampling, multi-structure data, top-$k$ list

\end{keyword}

\end{frontmatter}


\section{Introduction}
Robust model fitting techniques play an integral role in computer vision since the observations or measurements are frequently contaminated with 
outliers. Major applications include the estimation of various projective entities from multi-view data~\cite{multiview04} which often contain false 
correspondences. At the core of many robust techniques is random hypothesis generation, i.e., iteratively generate many hypotheses of the 
geometric model from randomly sampled minimal subsets of the data. The hypotheses are then scored according to a robust criterion (e.g., 
RANSAC~\cite{ransac}) or clustered (e.g., Mean Shift~\cite{meanshift}) to find the most promising model(s). Success rests upon retrieving an 
adequate number of \emph{all-inlier} minimal subsets which may require a large enough number of sampling steps.

This paper addresses two major issues affecting the current paradigm of robust estimation. The first is that hypothesis generation tends to be time 
consuming for heavily contaminated data. Previous methods attempted to improve sampling efficiency by guiding the sampling such that the 
probability of selecting all-inlier minimal subsets is increased. These methods often depend on assumptions or domain knowledge of the data, e.g., 
inliers have higher keypoint matching scores~\cite{gmlesac,prosac} or are correspondences that respect local geometry patterns~\cite{scramsac}. 
Most methods, however, are not optimized for data with \emph{multiple instances} (or \emph{structures}~\cite{stewart99}) of the geometric model. 
This is because they sample based on estimated inlier probabilities alone while ignoring the fact that only inliers from the \emph{same} structure 
should be included in the same minimal subset. Such methods may inefficiently generate a large number of samples before obtaining an all-inlier 
minimal subset for each genuine structure in the data.

\begin{figure}[t]
\centering
\subfigure[]{
\includegraphics[width=.36\linewidth]{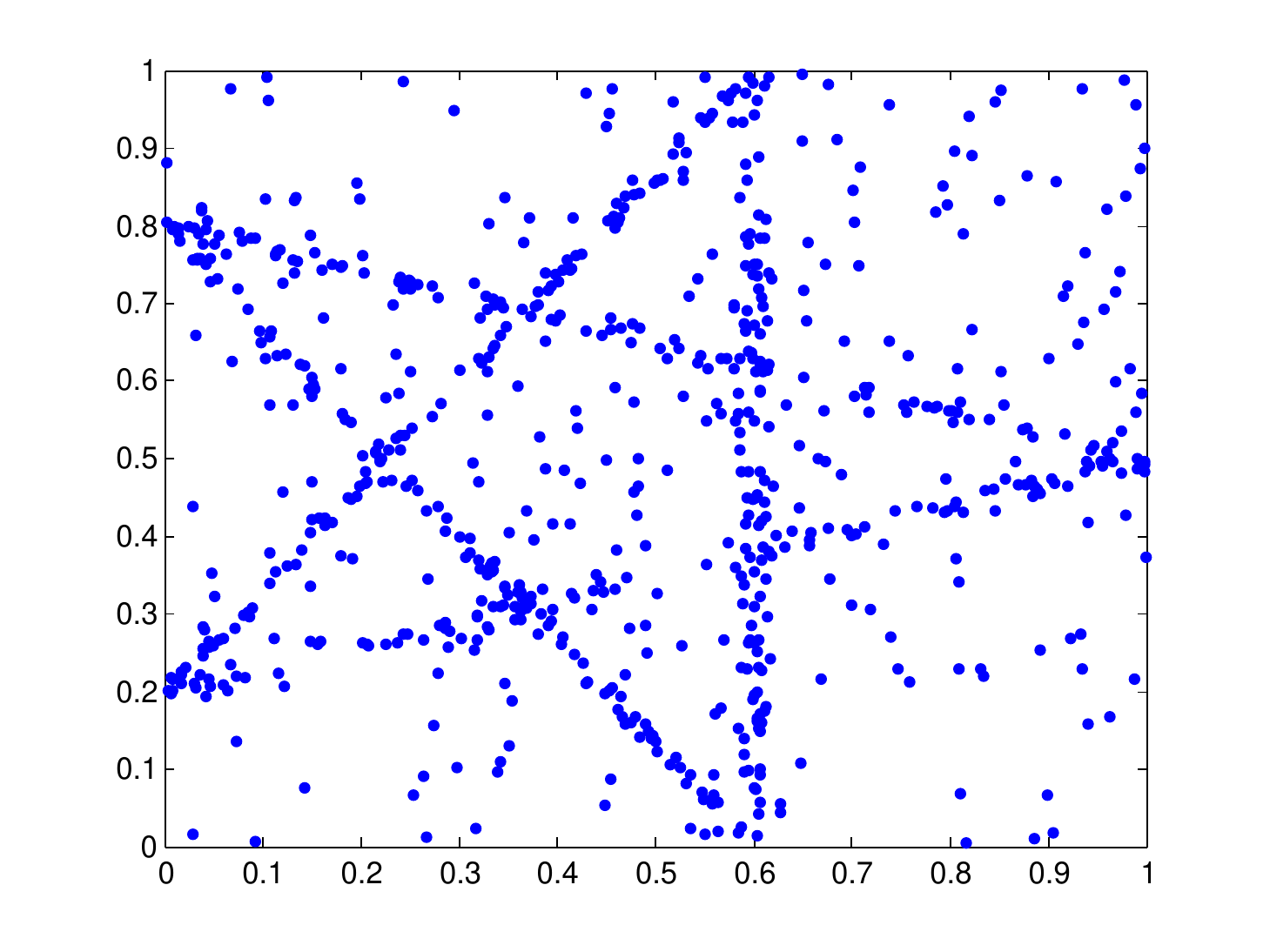}
\label{fig:star}
}
\hspace{-1.2cm}
\subfigure[]{
\includegraphics[width=.36\linewidth]{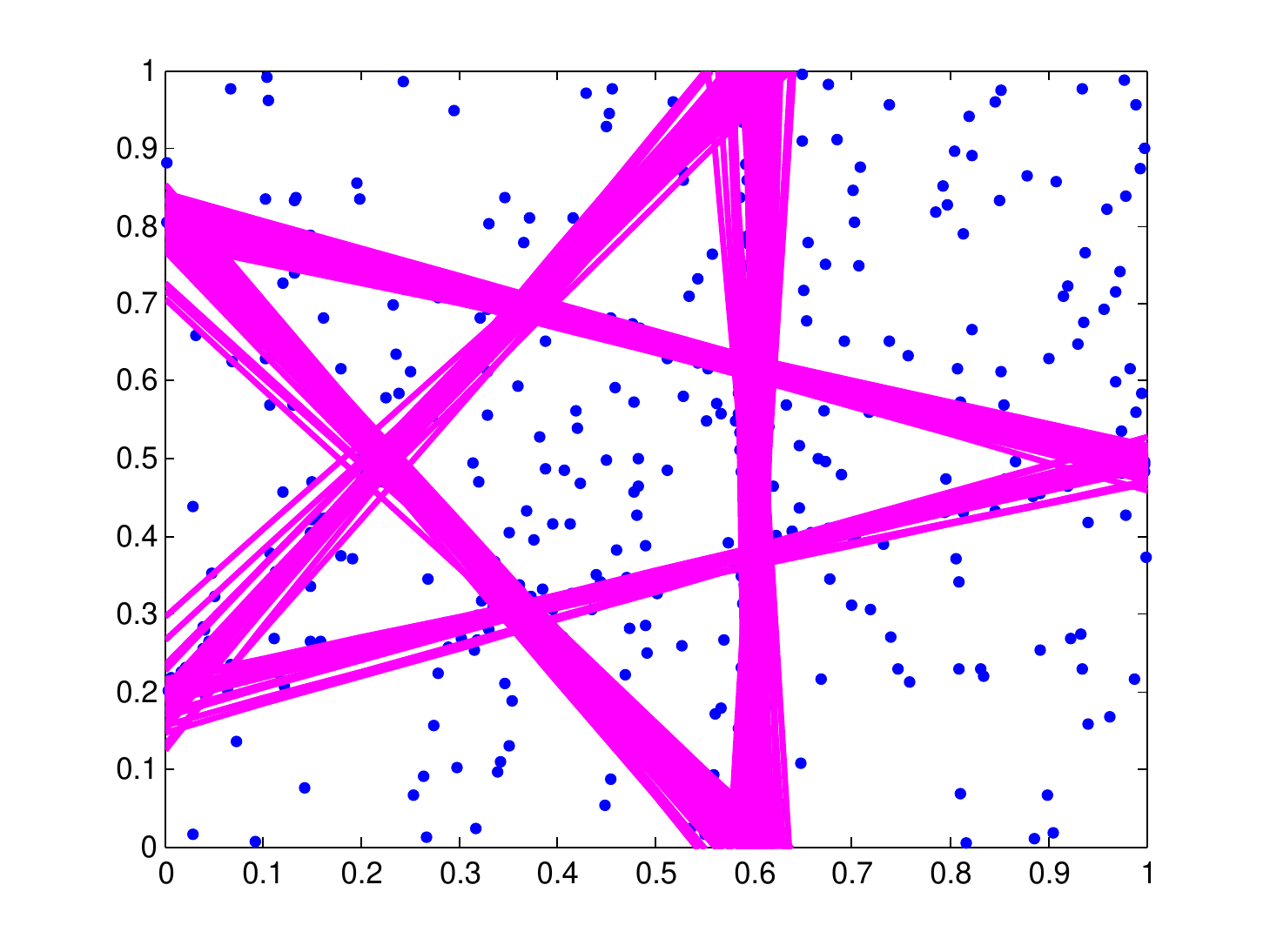}
\label{fig:stargoodh}
}\hspace{-1.1cm}
\subfigure[]{
\includegraphics[width=.36\linewidth]{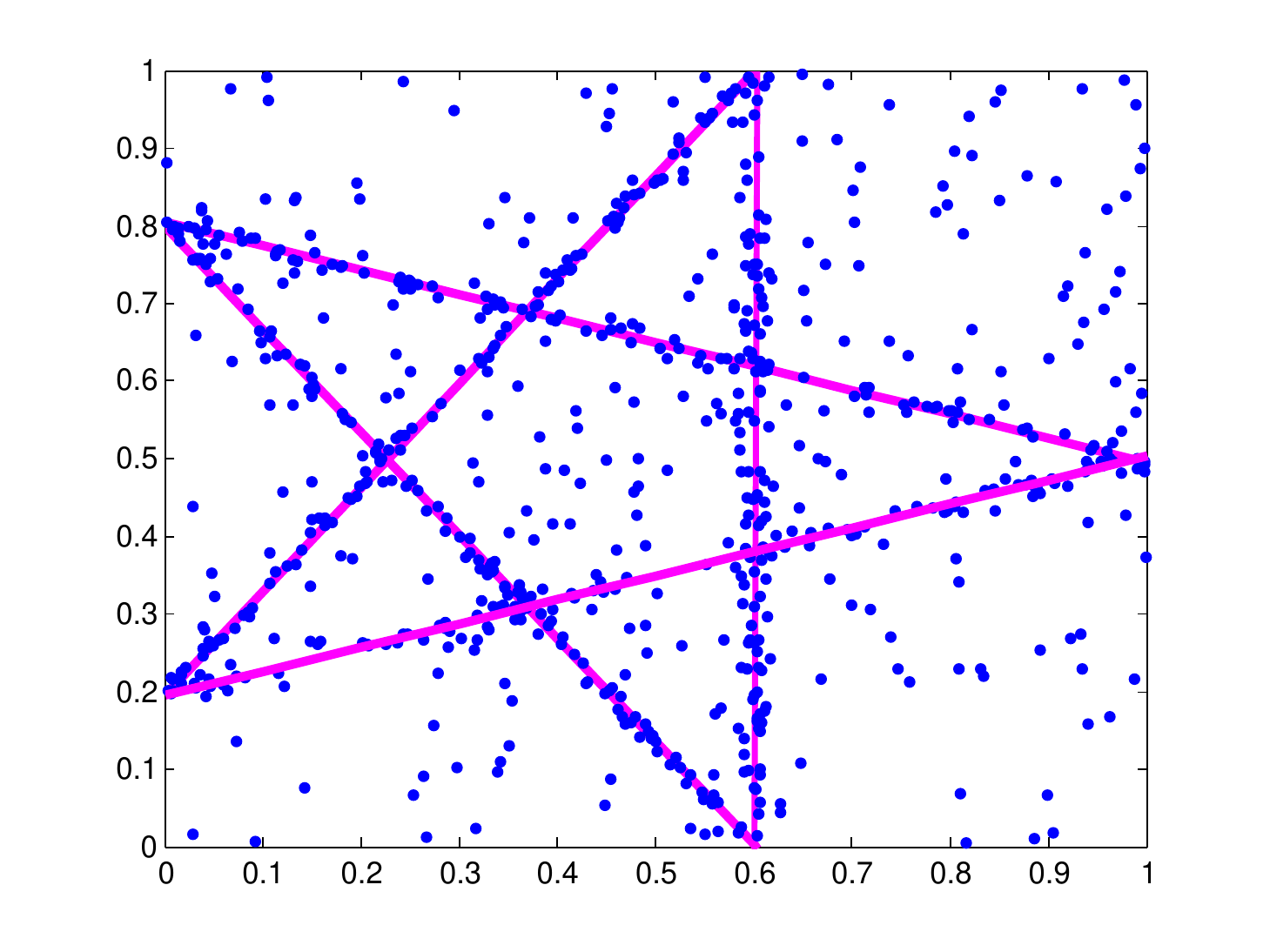} 
\label{fig:starfinal}
}
\caption{(a) Input data with 5 structures (lines) with 100 points per structure and 250 gross outliers. The inlier scale is 0.01. (b) 500 hypotheses are 
generated with the proposed \emph{multi-structure} guided sampling scheme and \emph{simultaneous} hypothesis filtering, producing 180 good 
hypotheses as shown in the figure. (c) Simple agglomerative clustering of the remaining 180 hypotheses gives the final fitting result.} 
\label{fig:summary}
\end{figure}

The second crucial issue is the lack of a principled approach to fit the multiple structures in the data. Many previous works~\cite{seqransac,proximity} 
simply apply RANSAC sequentially, i.e., fit one structure, remove corresponding inliers, then repeat. This is risky because inaccuracies in the initial fits 
will be amplified in the subsequent fits~\cite{multiransac}. Moreover, finding a stopping criterion for sequential fitting that accurately reflects the true 
number of structures is non-trivial. Methods based on clustering~\cite{jlinkage} or mode detection~\cite{meanshift,rht} given the generated 
hypotheses are not affected by the dangers of sequential fitting. However, if there are insufficient hypotheses corresponding to the true structures, 
the genuine clusters will easily be overwhelmed by the irrelevant hypotheses. Consequently, these methods often miss the true structures or find 
spurious structures.

The inability to retrieve ``good" hypotheses at sufficiently large quantities represents the fundamental obstacle to the satisfactory performance of 
previous methods. To address this limitation, we propose a novel hypothesis sampling scheme based on incremental computation of distances between 
\emph{partial rankings} or \emph{top-$k$ lists}~\cite{topklist} derived from residual sorting information. Our approach enhances hypothesis 
generation in two ways: (1) The computed distances guide the sampling such that inliers from a \emph{single} coherent structure are more likely to be 
simultaneously selected. This dramatically improves the chances of hitting all-inlier minimal subsets for \emph{each} structure in the data. (2) The 
qualities of the generated hypotheses are evaluated based on the computed distances. This permits an on-the-fly filtering scheme to reject ``bad'' 
hypotheses. The outcome is a set of only the most promising hypotheses which facilitate a simple agglomerative clustering step to fit all the genuine 
structures in the data. Fig.~\ref{fig:summary} summarizes the proposed approach.

The rest of the paper is organized as follows: Sec.~\ref{sec:gs} describes how to derive data similarities from residual sorting information by 
comparing top-$k$ lists. Sec.~\ref{sec:gshf} describes our guided sampling scheme with simultaneous hypothesis filtering and incremental 
computations of distances between top-$k$ lists. Sec.~\ref{sec:mms} describes how multi-structure fitting can be done by a simple agglomerative 
clustering on the promising hypotheses returned by our method. Sec.~\ref{sec:results} presents results on synthetic and real data which validate our approach. Finally, we draw conclusions in Sec.~\ref{sec:conclusion}.


\section{Data Similarity by Comparing Top-$k$ Lists}
\label{sec:gs}
A key ingredient of our guided sampling scheme is a data similarity measure. This section describes how to derive such a measure from residual sorting 
information.

\subsection{Top-$k$ Lists from Residual Sorting Information}
\label{sec:constructtopklist}
We measure the similarity between two input data based on the idea that if they are inliers from the same structure, then their preferences to the 
hypotheses as measured by residuals will be similar.  Such preferences can be effectively captured by lists of ranked residuals.

Let $X=\{x_{i}\}_{i=1}^{N}$ be a set of $N$ input data and $\Theta=\{\theta_{j}\}_{j=1}^{M}$ a set of $M$ hypotheses, where each hypothesis $\theta_{j}$ is fitted from a minimal subset of $p$ data (e.g., $p$=2 for line fitting). For each datum $x_{i}$, we compute its absolute residual as measured to $M$ hypotheses to form a residual vector 
\begin{align}
r_i~=~[r_{1}^{(i)},r_{2}^{(i)},\cdots,r_{M}^{(i)}]. 
\end{align}
We sort the elements in $r_{i}$ to obtain a sorted residual vector 
\begin{align}
	\tilde{r}_{i}~=~[r_{\lambda_{1}^{(i)}}^{(i)},\cdots,r_{\lambda_{M}^{(i)}}^{(i)}]
\end{align}
such that $r_{\lambda_{1}^{(i)}}^{(i)}\leq\cdots\leq r_{\lambda_{M}^{(i)}}^{(i)}$. The permutation $[\lambda_{1}^{(i)},\cdots,\lambda_{M}^{(i)}]$ encapsulate the data preference of $x_i$ to the hypotheses, i.e.,  $x_{i}$ is more likely to be an inlier to the hypotheses which have higher rank.

The top-$k$ list of data $x_{i}$ is defined as the first $k$ elements in the permutation $[\lambda_{1}^{(i)},\cdots,\lambda_{M}^{(i)}]$ , i.e.,
\begin{align}
  \tau_{i}~=~[\lambda_{1}^{(i)},\cdots,\lambda_{k}^{(i)}].
  \label{eq:topklist}
\end{align}
The top-$k$ list $\tau_{i}$ essentially gives the top-$k$ hypotheses preferred by $x_i$.

\begin{figure}[tb]
\centering
\includegraphics[width=.5\linewidth]{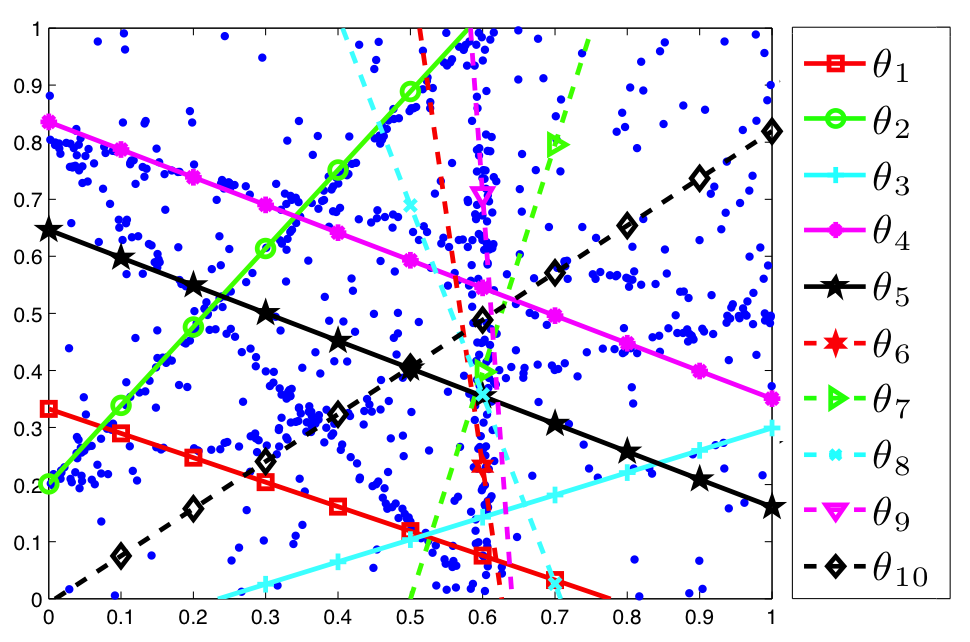}\hspace{0.3cm} \vspace{0.5cm}
\includegraphics[width=.425\linewidth]{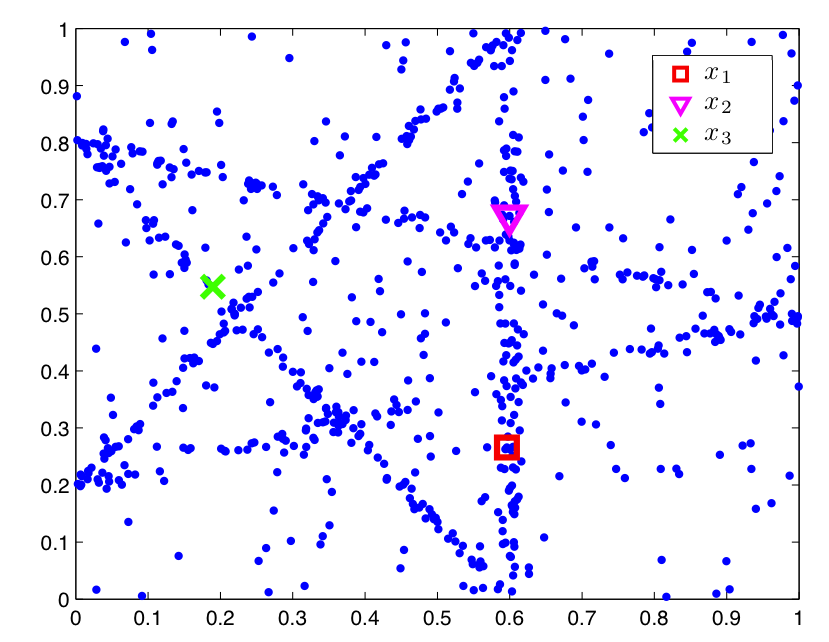} 
\includegraphics[width=.6\linewidth]{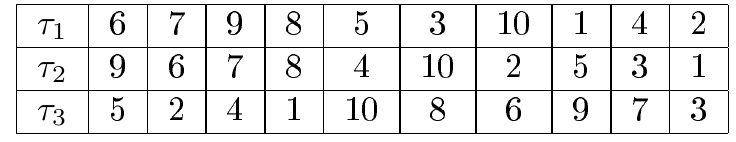}
\caption{Top Left: 10 hypotheses are randomly generated using input data in Fig.~\ref{fig:star}. Top Right: 3 data are selected ($x_1$ and $x_2$ are inliers from the same structure; $x_3$ is an inlier from different structure). Bottom: top-10 hypotheses corresponds to the selected data. (Best viewed in color) } 
\label{fig:motivation}
\end{figure}

Fig~\ref{fig:motivation} illustrate our idea to measure the data similarity using data preferences to the hypotheses. For data $x_1$ and $x_2$, they are inliers from the same structure and their corresponding top-10 lists $\tau_1$ and $\tau_2$ are similar, e.g., hypotheses 6,7 and 8 are highly ranked. For two inliers $x_1$ and $x_3$ from different structure,  their corresponding top-10 lists $\tau_1$ and $\tau_3$ are inconsistent, e.g., hypotheses 2 and 4 are highly ranked by $x_3$ but not by $x_1$.

\subsection{The Spearman Footrule Distance}
\label{sec:comparetopklist}
Given the top-$k$ lists, we measure their similarity using the Spearman Footrule (SF) distance \cite{topklist}. Let $\tau$ be a top-$k$ list and 
$D_\tau$ be a set of elements contained in $\tau$. Denote the position of the element $m \in D_{\tau}$ in $\tau$ by $\tau{(m)}$. The SF distance 
between two top-$k$ lists $\tau_{i}$ and $\tau_{i^\prime}$ is defined as
\begin{align}
  F^{(\ell)}{(\tau_{i},\tau_{i^\prime})}~=~\sum_{m \in D_{\tau_{i}}\cup
    D_{\tau_{i^\prime}}}\left|\tau_{i}^{\prime}(m)-\tau_{i^\prime}^{\prime}(m)\right|,
  \label{eq:sflp}
\end{align}
where $\ell>0$ is the so-called location parameter (often set to $k+1$), $\tau^{\prime}_{i}(m)=\tau_{i}(m)$ if $m \in D_{\tau_{i}}$; otherwise 
$\tau^{\prime}_{i}(m)=\ell$, and $\tau^{\prime}_{i^\prime}$ is similarly obtained from $\tau_{i^\prime}$.

\subsection{Measuring Similarity between Data}
\label{sec:similaritybetweendata}
To measure the similarity between two data, we use the SF distance (Eq.~\ref{eq:sflp}) between their corresponding top-$k$ lists. The similarity 
value between two data $x_{i}$ and $x_{i^\prime}$ is defined as
\begin{align}
d{(\tau_{i},\tau_{i^\prime})} = 1-\frac{1}{k\times\ell}F^{(\ell)}{(\tau_{i},\tau_{i^\prime})}.
\label{eq:K}
\end{align}
Note that we normalize $F^{(\ell)}{(\tau_{i},\tau_{i^\prime})}$ such that $d{(\tau_{i},\tau_{i^\prime})}$ is between 0 (dissimilar) and 1(identical). By comparing 
the top-$k$ lists between all data, we obtain a $N\times N$ similarity matrix $K$ with 
\begin{align}
K(i,i^\prime)=d{(\tau_{i},\tau_{i^\prime})},
\label{eq:Kmatrix}
\end{align}
where $K(i,i^\prime)$ denotes the element at its $i$-th row and $i^\prime$-th column. 

\begin{figure}[tb]
\centering
\subfigure[]{
\includegraphics[width=.33\linewidth]{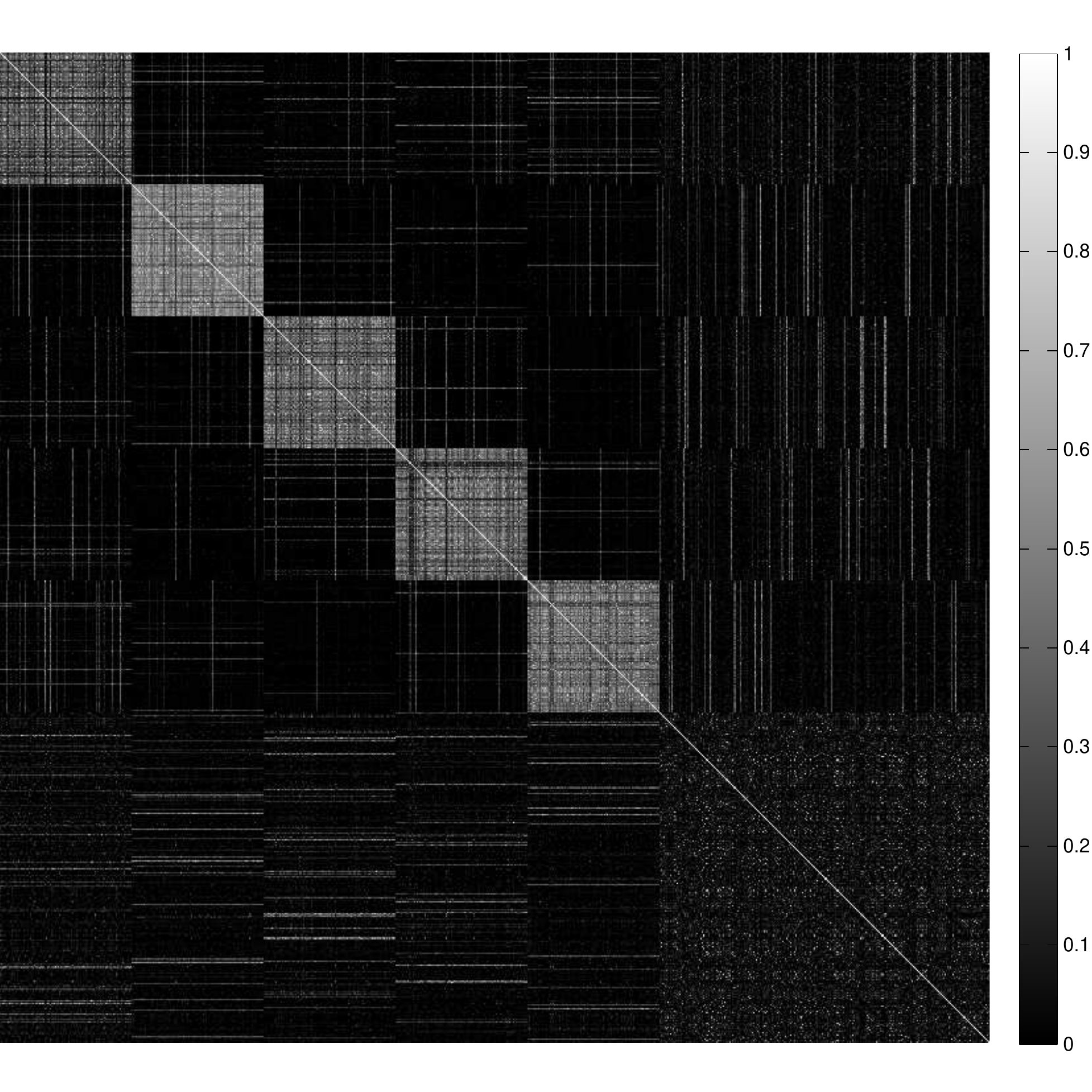}
\label{fig:stark}
}\hspace{0.5cm}
\subfigure[]{
\includegraphics[width=.55\linewidth]{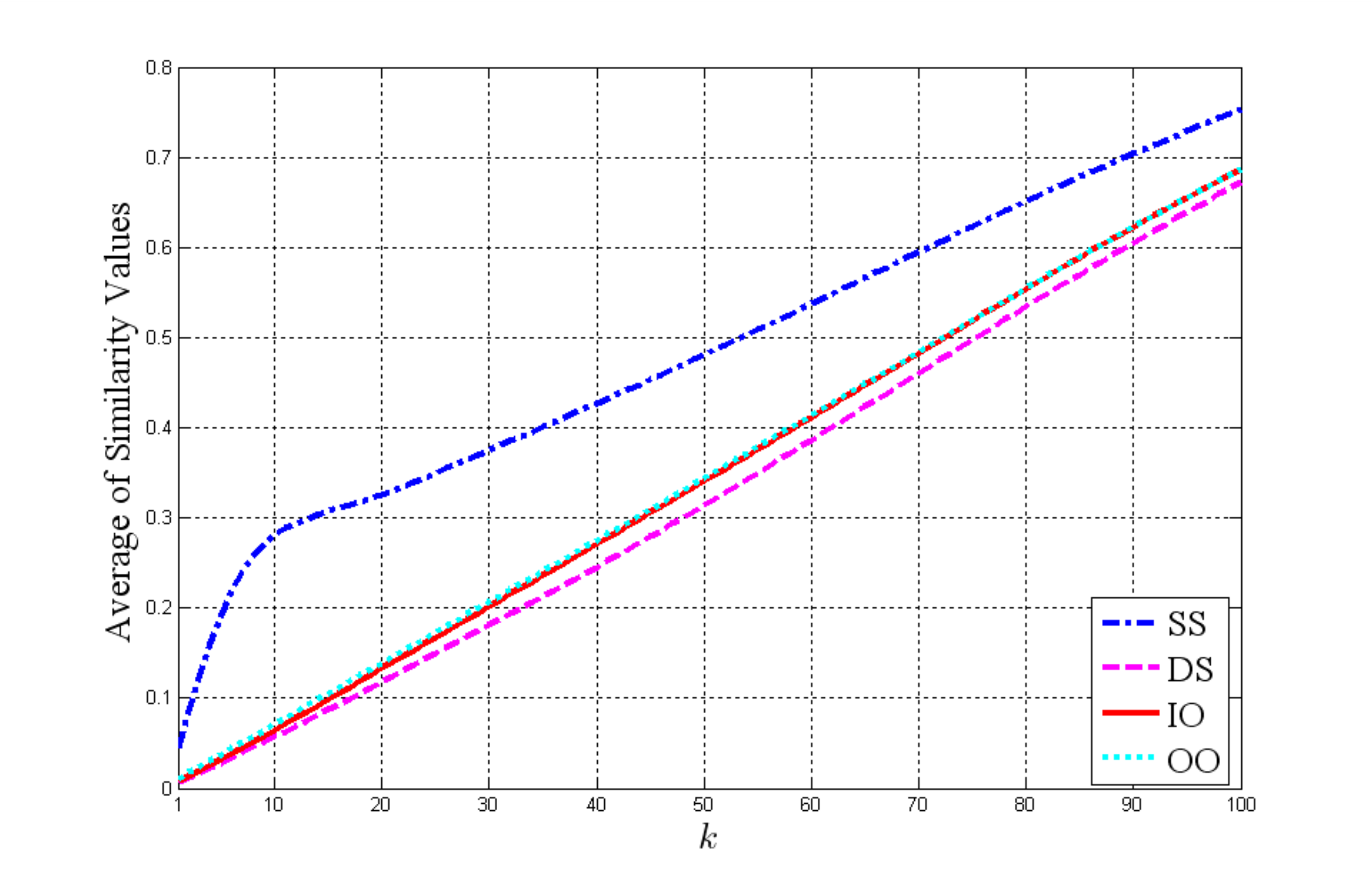}
\label{fig:star_ss_ds}
}
\caption{(a) Similarity matrix $K$ for data shown in Fig.~\ref{fig:star} (data is arranged according to structure membership for representation only) 
(b) Under various k, the average of similarity values between two inliers from the same structure (SS), two inliers from different structures (DS), an inlier and a gross outlier (IO), two gross outliers (OO).} 
\label{fig:K}
\end{figure}

Fig.~\ref{fig:stark} shows an example of $K$, which is generated from the input data shown in Fig.~\ref{fig:star}. The evident block structures in $K$ correspond to the 5 lines in Fig.~\ref{fig:star}. As shown in Fig.~\ref{fig:star_ss_ds}, across various $k$, the average of similarity values between two inliers from the same structure (SS) is higher than that from different structure (DS). It supports the observation from our given example (Fig.~\ref{fig:motivation}) that inliers from the same structure gives consistent preferences to the hypotheses. From Fig.~\ref{fig:star_ss_ds}, it also shows that the inconsistent preferences to hypotheses occur between an inlier/a gross outlier(IO) and two gross outliers (OO), hence, their corresponding averages of similarity values are smaller. 

\section{Guided Sampling with Hypothesis Filtering}
\label{sec:gshf}
This section describes our guided sampling scheme which involves a simultaneous hypothesis filtering scheme. We also provide an efficient incremental update for computing the sampling weights. Furthermore, we provide an improvement on our guided sampling scheme based on the result from our hypothesis filtering scheme.

\subsection{Guided Sampling}
\label{sec:guidedsampling}
\begin{figure}[tb]
\centering
\includegraphics[width=0.95\linewidth]{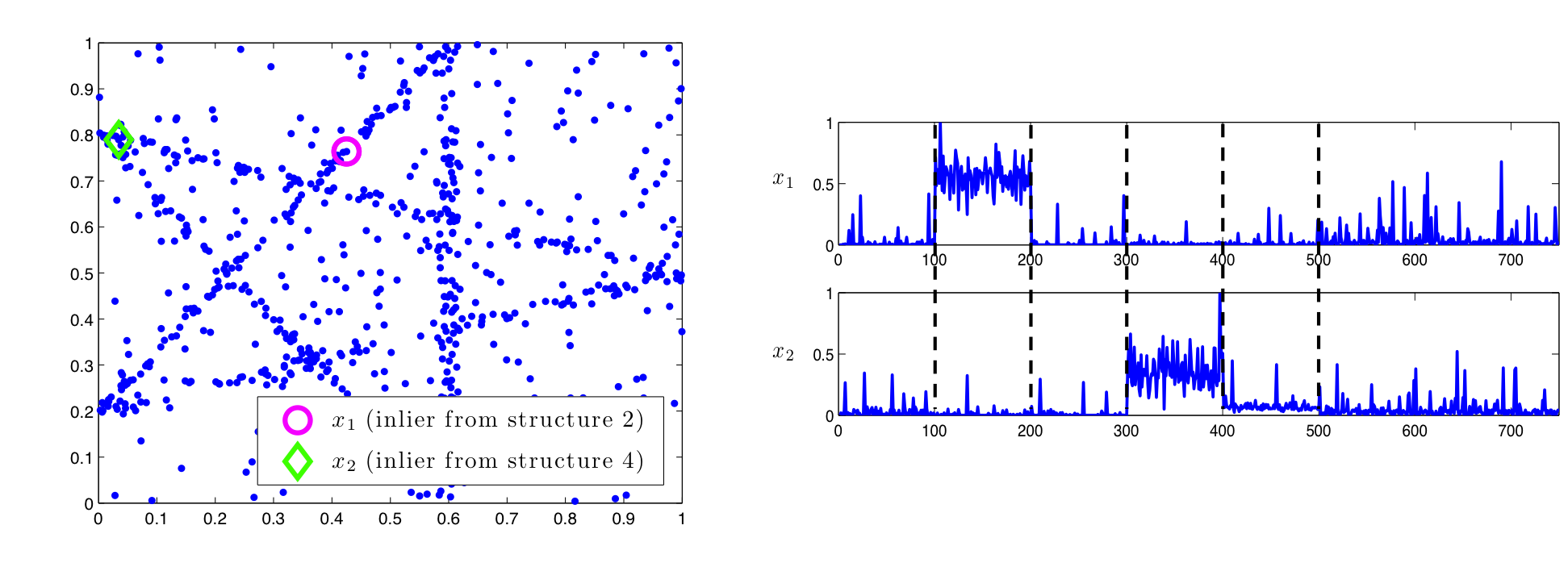}
\caption{Example of sampling weights. Left: two inliers from different structure. Right: their corresponding sampling weights for selecting the next datum. The data is arranged based on the structure membership (denoted by vertical dashed lines) for representation only.}
\label{fig:showweight}
\end{figure}

We use the similarity matrix $K$ (Eq.~\ref{eq:Kmatrix}) to sample data in a guided fashion. Let $Q=\{s_{u}\}_{u=1}^{p}$ be the indices of data in a 
minimal subset of size $p$, where $s_{u}$ are indexed by the order in which they are sampled. The first element $s_{1}$ in $Q$ is randomly selected 
from $X$. To sample the next element $s_{2}$, we use $K{(s_{1},:)}$ as the weight to guide the sampling, i.e., the similarity values of all input data with respect to $s_{1}$. We set $K{(s_{1},s_{1})}$ to 0 to avoid sampling the same data again. Fig.~\ref{fig:showweight} shows the example of such sampling weights. 

Suppose data $s_{1},\cdots,s_{u}$ have been selected, then the next datum $s_{u+1}$ is chosen conditionally on the selected data. Its sampling 
weight is defined as
\begin{align}
  K^\prime{(s_{1},:)}~\cdot~K^\prime{(s_{2},:)}~\cdot~\ldots~\cdot~K^\prime{(s_{u},:)},
  \label{eq:weght}
\end{align}
where $\cdot$ is the element-wise multiplication and $K^\prime{(s_{u},:)}$ is just $K{(s_{u},:)}$ with $K{(s_{u},s_{u})}=0$. Eq. \ref{eq:weght} 
means that in order to have higher  probabilities of being sampled, a datum need to be similar (measured by Eq. \ref{eq:K}) to all the data that have been selected into the minimal subset.

\subsection{Incremental Top-$k$ Lists Comparison}
\label{sec:incrementaltopk}
Our sampling method computes an update to the similarity matrix $K$ (Eq.~\ref{eq:Kmatrix}) once a block (of size $b$) of new hypotheses are generated. This 
involves comparing top-$k$ lists of ranked residuals. The computation of top-$k$ lists can be done efficiently via merge sort. However, comparing 
top-$k$ lists between all data, i.e., constructing $K$, can be computationally expensive. Here we provide efficient incremental updates for $K$ that 
can substantially accelerate the computation.

As proved in \cite{topklist}, the SF distance (Eq.~\ref{eq:sflp}) can be equivalently computed as 
\begin{align}
F^{(\ell)}{(\tau_{i},\tau_{j})} = 2(k-\left|Z\right|)\ell+\!\sum_{m\in Z}\left|\tau_{i}(m)-\tau_{j}(m)\right|-\sum_{m \in S}\tau_{i}(m)-\sum_{m\in 
T}\tau_{j}(m)\mbox{,}
\label{eq:long}
\end{align}
where $Z=D_{\tau_{i}}\cap D_{\tau_{j}}$, $S=D_{\tau_{i}}\backslash D_{\tau_{j}}$ and $T=D_{\tau_{j}}\backslash D_{\tau_{i}}$. In fact, $S$ is 
simply the elements in $D_{\tau_{i}}$ but not in $Z$, i.e., $S=D_{\tau_{i}}\backslash Z$, similarly for $T$. Hence, we have
\begin{align}
\sum_{m\in S}\tau_{i}(m) = \sum_{m=1}^{k}m-\sum_{m\in Z}\tau_{i}(m) = \frac{1}{2}k(k+1)-\sum_{m\in Z}\tau_{i}(m),
\label{eq:s}
\end{align}
similarly for $\sum_{m\in T}\tau_{j}(m)$. By setting $\ell=k+1$ and using Eq.~\ref{eq:s}, we can rewrite Eq.~\ref{eq:long} to be in terms of $Z$ 
only,
\begin{align}
F^{(k+1)}{(\tau_{i},\tau_{j})}=(k+1)(k-2\left|Z\right|)\!+\!\!\!\!\sum_{m\in Z}\!\!\!\left(\left|\tau_{i}(m)-\tau_{j}(m)\right|+\tau_{i}(m)+\tau_{j}
(m)\right).
\label{eq:modifiedlong}
\end{align}

Let $A$ and $B$ be two $N\times N$ symmetric matrices and set the elements at the $i$-th row and the $j$-th column of $A$ 
and $B$ to
\begin{align}
A(i,j)=\left|Z\right|\mbox{~and~}B(i,j)=\sum_{m\in Z}\left( \left|\tau_{i}{(m)}-\tau_{j}{(m)}\right|+\tau_{i}{(m)}+\tau_{j}{(m)}\right).
\label{eq:AB}
\end{align}
From Equations~\ref{eq:K}, \ref{eq:Kmatrix}, and \ref{eq:modifiedlong}, the similarity matrix $K$ can be constructed by 
\begin{align}
K = 1-\frac{1}{k}(k-2A)-\frac{1}{k(k+1)}B,   
\label{eq:incrementalk}
\end{align}
where $I_N$ is an $N\times N$ identity matrix. Observe from Eq.~\ref{eq:AB} that the matrices $A$ and $B$ can be efficiently updated by keeping 
track of the elements that move into or out of $Z$. This information is readily available from the merge sort. Once $A$ and $B$ are updated, 
$K$ can be updated via Eq.~\ref{eq:incrementalk}.
%

\subsection{Simultaneous Hypothesis Filtering}
\label{sec:sim}
During sampling, we want to simultaneously filter hypotheses such that only a small but very promising subset remains. We make use the data preferences and hypothesis preferences to identify the good hypotheses.

Similar to the definition of top-$k$ hypotheses in Eq.~\ref{eq:topklist}, we define the top-$h$ data of a hypotheses by sorting the residual information. For each hypothesis $\theta_{j}$, we sort its absolute residual $r_{j}=[r_{1}^{(j)},r_{2}^{(j)},\cdots,r_{N}^{(j)}]$ as measured to $N$ data and its sorted residual vector $\tilde{r}_{j}=[r_{\pi_{1}^{(j)}}^{(j)},\cdots,r_{\pi_{N}^{(j)}}^{(j)}]$ such that $r_{\pi_{1}^{(j)}}^{(j)}\leq\cdots\leq r_{\pi_{N}^{(j)}}^{(j)}$. The top-$h$ data of hypothesis $\theta_{j}$ is defined as the first $h$ elements in the permutation $[\pi_{1}^{(j)},\cdots,
\pi_{N}^{(j)}]$, i.e.,
\begin{align}
  \sigma_{j}= [\pi_{1}^{(j)},\cdots,\pi_{h}^{(j)}].
  \label{eq:tophdata}
\end{align}
The top-$h$ data $\sigma_{j}$ gives the $h$ data preferred by the hypothesis $\sigma_{j}$ to be its inliers, i.e., the higher $x_i$ is ranked, the more likely $x_i$ is the inlier to it. The value of $h$ is conservatively set to $0.1$ in all experiments. The assumption is that at least 10\% of data are inliers.

\begin{figure}[tb]
\centering
\subfigure[]{
\includegraphics[width=.5\linewidth]{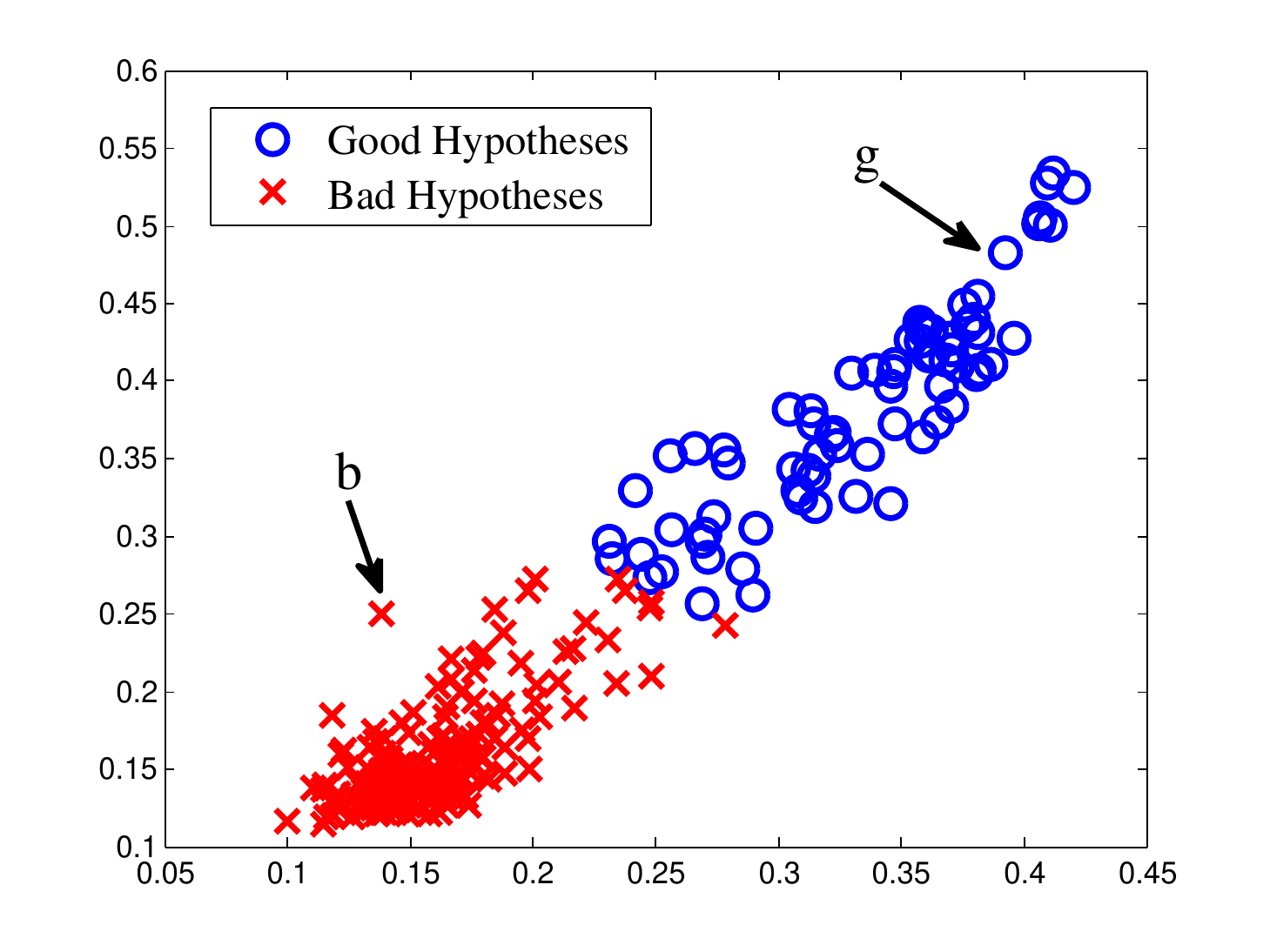}
\label{fig:featurespace}
}\hspace{-1cm}
\subfigure[]{
\includegraphics[width=.5\linewidth]{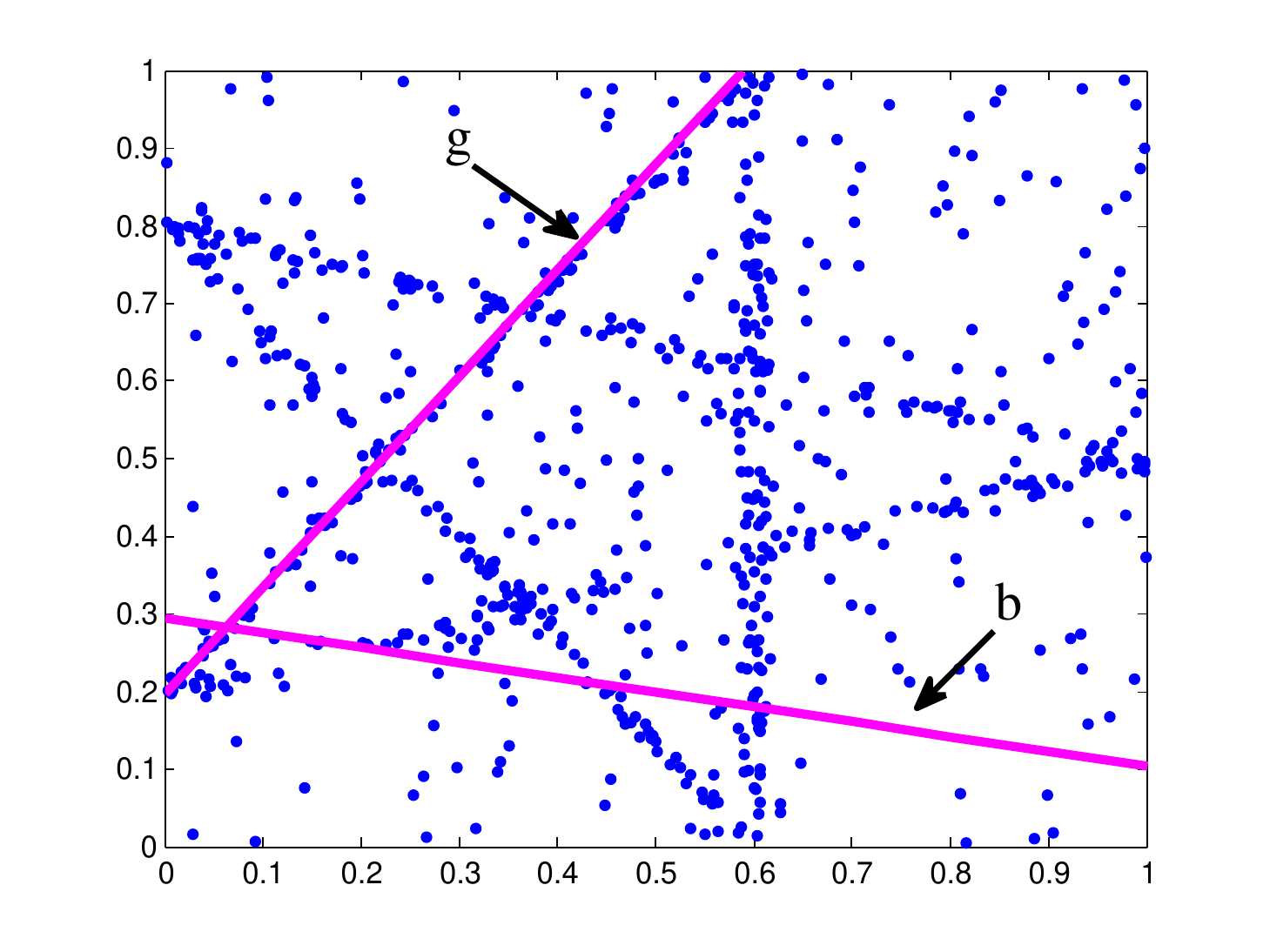}
\label{fig:bgexample}
}
\caption{Hypothesis filteirng: (Left) Feature space where $x$ axis is $f^{(1)}_{m}$ and $y$ axis is $f^{(2)}_{m}$ (Best view in color). (b) An example of ``good" (denoted 
``g") and ``bad" (denoted ``b") hypotheses.} 
\label{fig:featureexample}
\end{figure}

For each hypothesis $\theta_{j}$, we construct the feature vector using the data and hypothesis preferences.
\begin{align}
f_j = \left[{f^{(1)}_j,~f^{(2)}_j}\right] = \left[\frac{\sum_{(i,i^\prime)\in E}K(i,i^\prime)}{\left|E\right|},~\frac{\sum_{(i,i^\prime)\in \Sigma}K(i,i^\prime)}{\left|\Sigma\right|}\right],
\label{eq:feature}
\end{align}
where $E=\{(i,i^\prime)|i\neq i^\prime\mbox{~and~}x_i,x_{i^\prime}\in\Omega_j\}$ with $\Omega_j = \{x_i\in X\mid m\in \tau_i\}$, $\Sigma =\{(i,i^\prime)|i\neq i^\prime\mbox{~and~}i,i^\prime\in\sigma_j\}$ and $K$ is the similarity matrix computed by Eq.~\ref{eq:incrementalk}. The set $\Omega_j$ contains all data that include the hypothesis $j$ in their top-$k$ lists. If the 
hypothesis $j$ is ``good", then $\Omega_j$ should contain many inliers from a structure. Hence, $f^{(1)}_j$, the average of similarity values 
between all data in $\Omega_m$ should be high. Moreover, the top-$h$ data of hypotheses should contains data which are similar to each other (from the same structure), the average of similarity values between its top-$h$ data should be high, i.e., high $f^{(2)}_j$. Therefore, we want to find a set of hypotheses which have high value in both $f^{(1)}_m$ and $f^{(2)}_j$. To this end, we apply k-means on the feature vectors (Eq.~\ref{eq:feature}) to separate ``good" and ``bad" hypotheses. As illustrated in Fig.~\ref{fig:featureexample}(a), the cluster whose center has larger norm (circles) contains good hypotheses. We incrementally maintain a set of ``good" hypotheses as the guided sampling proceeds.

\subsection{Selection of the First Datum in Minimal Subset}
\label{sec:chooseseed}
The guided sampling scheme in Sec.~\ref{sec:guidedsampling} can be further improved by the result from the the hypothesis filtering scheme. 

In Sec.~\ref{sec:guidedsampling}, the first element in the minimal subset is selected randomly. The subsequent sampling the data of the minimal subset becomes unprofitable if the first element is a gross outlier. 
Here, we can make use the result from hypothesis filtering to increase the probability of obtaining an inlier. Note that if a hypothesis is ``good'', its minimal subset should mainly contain inliers. Hence, we simply select the first element from the minimal subsets of the ``good" hypotheses.

\section{Multi-Structure Fitting}
\label{sec:mms}  
By leveraging the simultaneous hypothesis filtering, a set of ``good" hypotheses is immediately available once the sampling is done. This set of ``good" hypotheses allows us to easily cluster the hypotheses using a simple agglomerative clustering method.

We use the agglomerative clustering (See~\cite{hastie2005elements} for a detailed description) to cluster the hypotheses. For each hypothesis $\theta_j$, we aggregate the similarity of its top-$h$ data as measured to $N$ data and represent each hypothesis $\theta_j$ by a $1\times N$ feature 
vector
\begin{align}
  \alpha_{j}=\sum_{i\in \sigma_j}K(i,:).
  \label{eq:clsuterid}
\end{align}
The distance between two hypotheses $\theta_j$ and $\theta_{j^\prime}$ is given by  
\begin{align}
  d(\theta_j, \theta_{j^\prime}) = \left\|\alpha_{j}-
      \alpha_{j^\prime}\right\|_{2},
  \label{eq:clsuterdistance}
\end{align}
where $\left\|\cdot\right\|_{2}$ denotes the $L_2$ norm. The $d(\theta_j, \theta_{j^\prime})$ is smaller if two hypotheses $\theta_j$ and $\theta_{j^\prime}$ is similar, i.e, explaining the same structure. Using this distance measure, the clustering is then performed through the standard agglomerative clustering mechanism.

Each cluster of hypotheses contains the hypotheses overlapping on the same structure. For each cluster of hypotheses, we can simply select the hypothesis with the minimum of sum of squared residuals over its top-$h$ data.

\section{Experiments}
\label{sec:results}
We test the proposed method on homography and fundamental matrix estimation using real data. To evaluate the efficiency of the proposed guided sampling scheme, we 
compare our method against 6 sampling techniques: Uniform random sampling in RANSAC (Random) \cite{ransac}, proximity sampling (Proximity) 
\cite{proximity,jlinkage}, LO-RANSAC(LRANSAC) \cite{loransac}, Guided-MLESAC (GMLESAC) \cite{gmlesac} and PROSAC \cite{prosac}. Our proposed method is denoted by ITKSF (Sec.~\ref{sec:guidedsampling}) and its extension is denoted by ITKSF-S (Sec.~\ref{sec:chooseseed}).

In all experiments, we fix $b = 100$ and $k = \lceil 0.1 \times t \rceil$ throughout, $b$ being the block size (cf Sec.~\ref{sec:incrementaltopk}) and $t$ being the number of hypotheses generated so far. All experiments are run on a machine with 2.53GHz Intel Core 2 Duo processor and 4GB RAM.

\subsection{Data Set: AdelaideRMF}
We created a data set for robust model fitting, called AdelaideRMF\footnote{{\scriptsize{AdelaideRMF is publicly available from \url{http://cs.adelaide.edu.au/~hwong/doku.php?id=data}}}}. It contains a collection of image pairs for homography and fundamental matrix estimation on single and multi-structure data. For each image pair, we use SIFT~\cite{sift} to obtain the keypoint correspondences and manually labelled each keypoint correspondence.

\subsection{Homography Estimation}
\begin{figure}[H]
\centering
\subfigure[Union]{
\includegraphics[width=.45\linewidth]{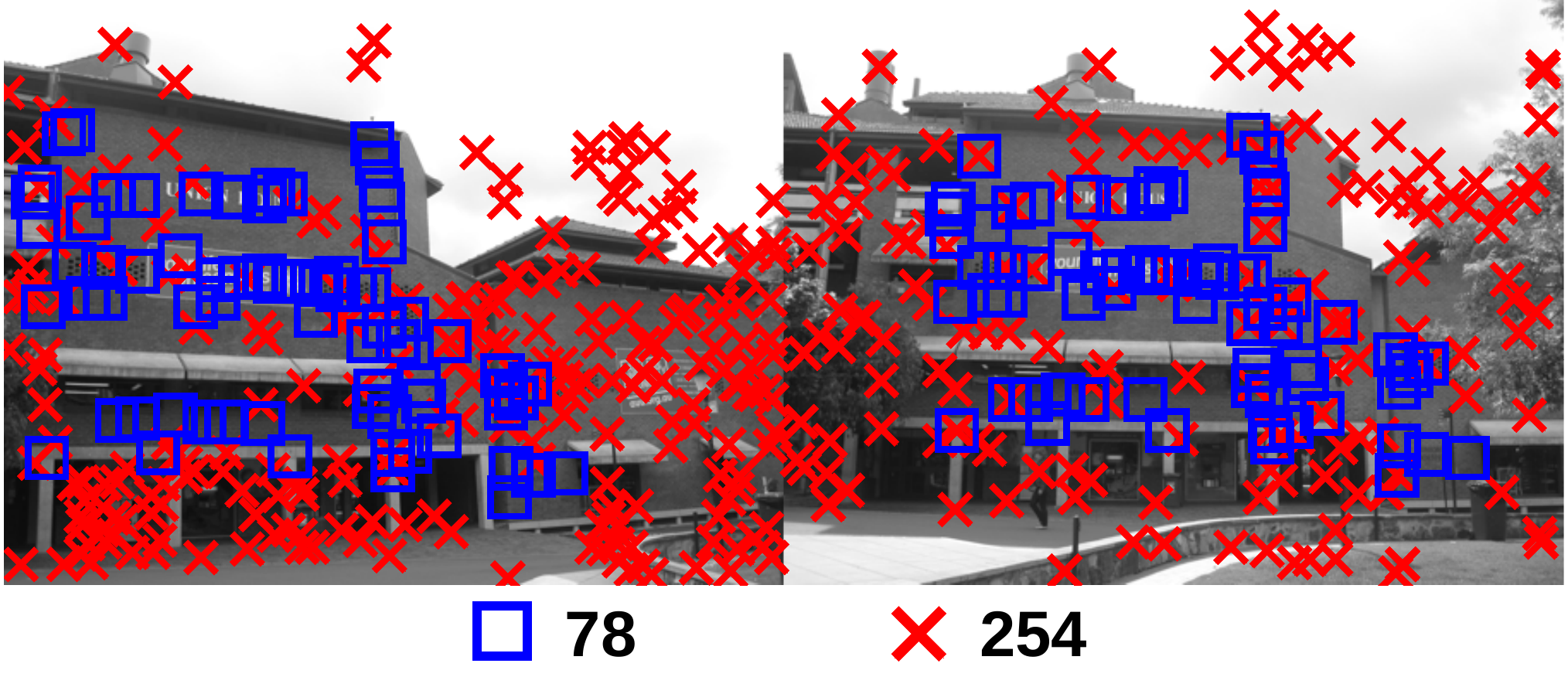}
\label{fig:union}
}

\subfigure[Hartley]{
\includegraphics[width=.45\linewidth]{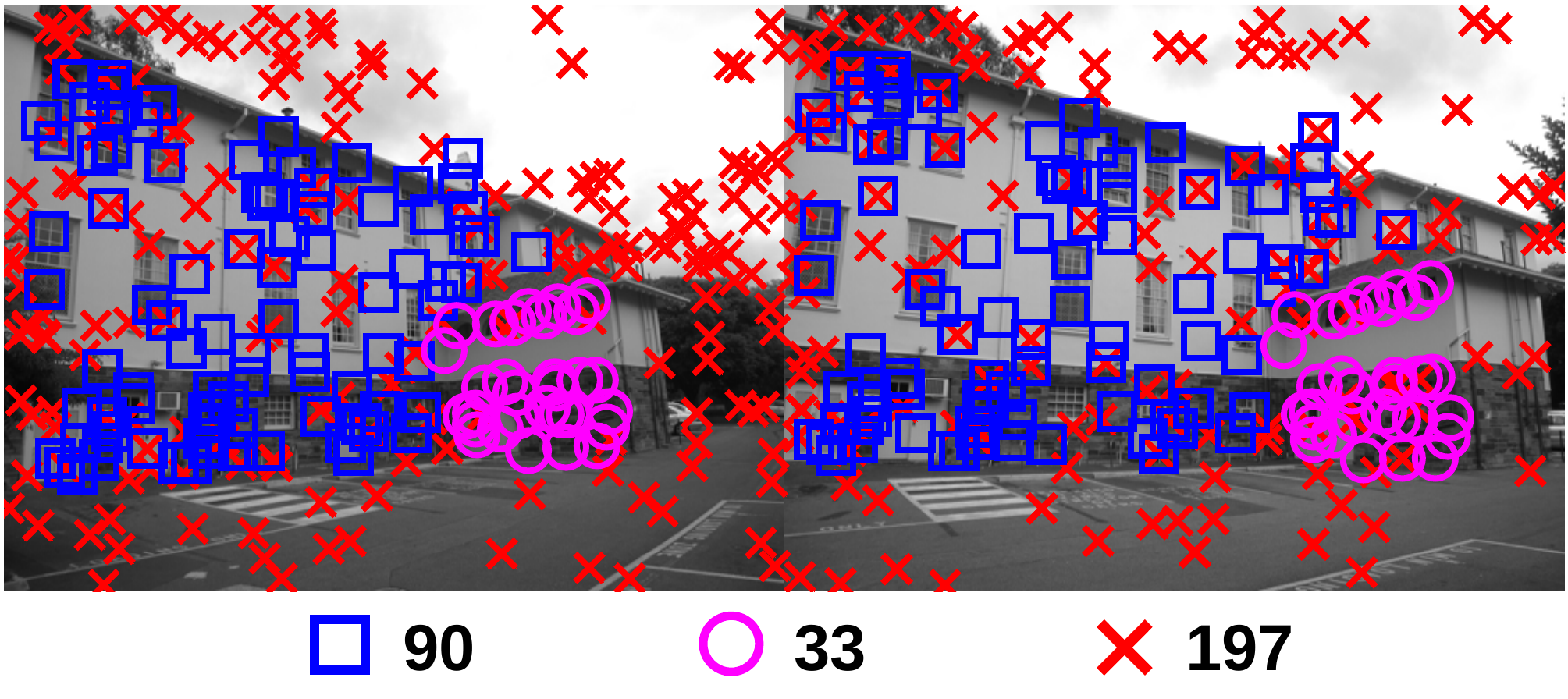}
\label{fig:hartley}
}
\subfigure[Symon]{
\includegraphics[width=.45\linewidth]{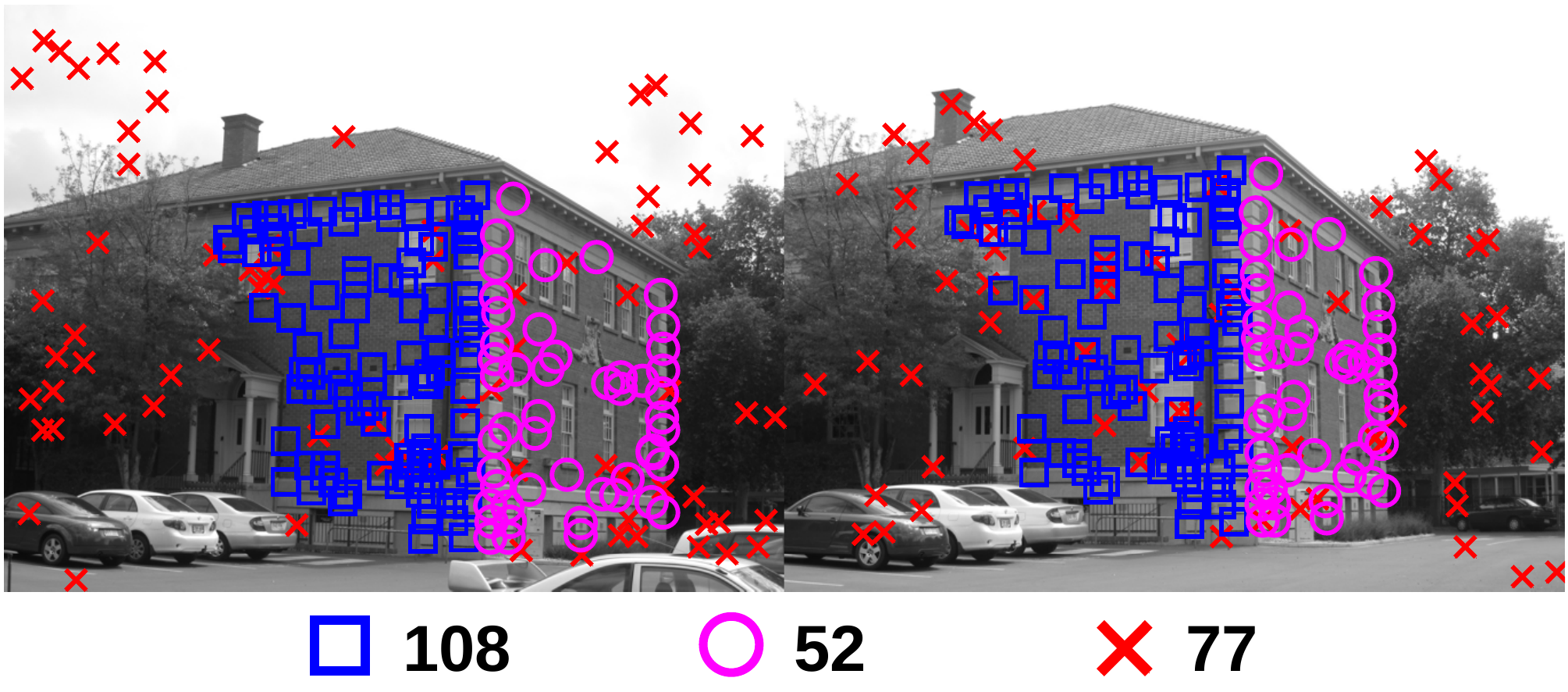}
\label{fig:ladysymon}
}
\subfigure[NEEM]{
\includegraphics[width=.45\linewidth]{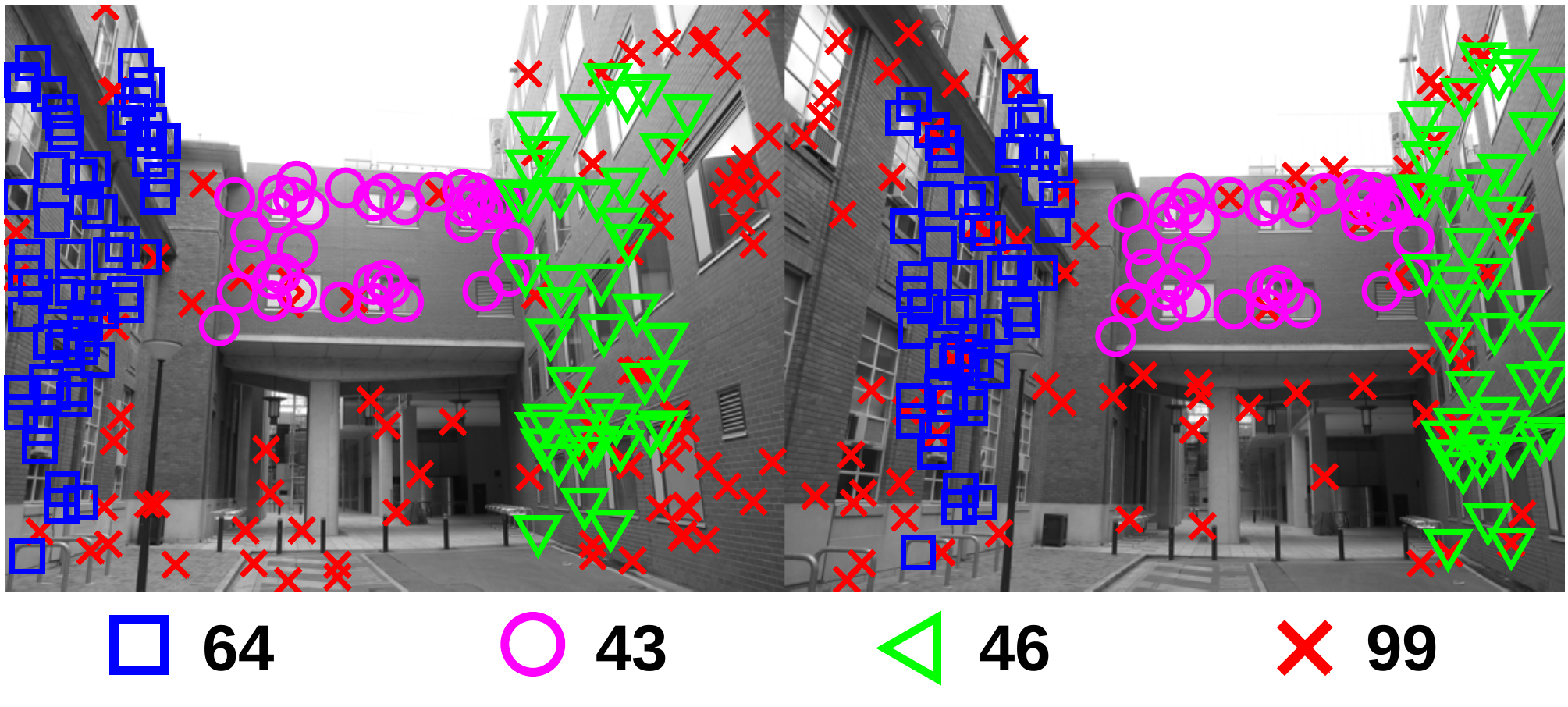}
\label{fig:neem}
}
\subfigure[Johnson]{
\includegraphics[width=.45\linewidth]{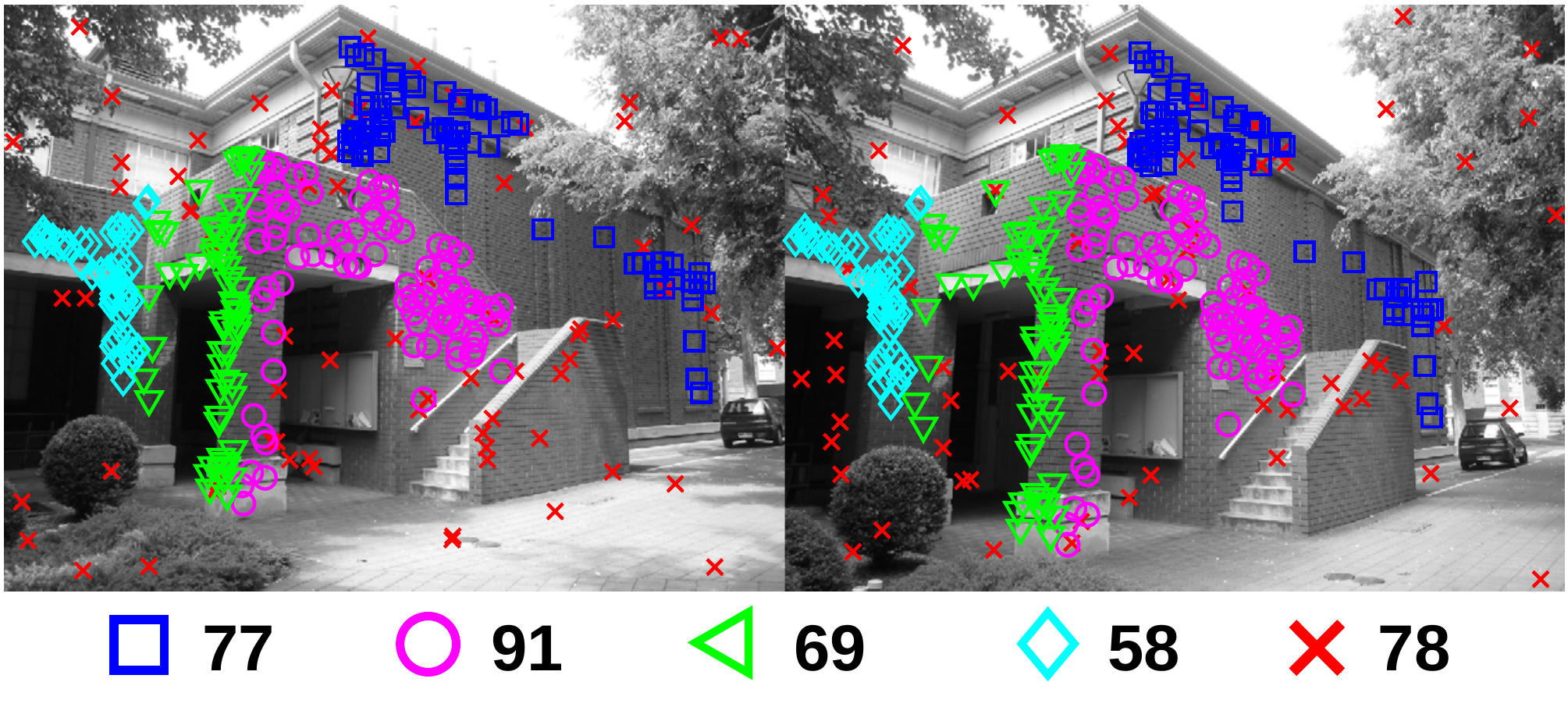}
\label{fig:johnsona}
}
\caption{Data for homography estimation. Red crosses indicate the gross outliers and other colored markers indicate the structure membership. (Best viewed in color)} 
\label{fig:homodata}
\end{figure}

This experiment involves estimating multiple planar homographies. The data used for this experiment is shown in ~Fig.~\ref{fig:homodata}. We use 4 correspondences to estimate a homography using Direct Linear Transformation \cite{multiview04}. Each method is given 50 random runs, each for 5 CPU seconds.

Table~\ref{tab:homotable} shows the performance of guided sampling methods and also the hypothesis filtering result from our proposed method.

\begin{table}[H]
  \centering
\caption{Performance of guided sampling methods in homography estimation. We record the number of hypotheses generated (M),the number of all-inlier minimal subset found on each structure (Structures) and the percentage of all-inlier minimal subsets (IS) found within the time budget. We also record the CPU time to hit at least one all-inlier minimal subset on each structure (HIT) which is penalized by the time budget if a method fails. For ITKSF and ITKSF-S, the hypothesis filtering result is shown in bracket. The results represent the average over 50 runs with the best result boldfaced.}
\label{tab:homotable}
{\fontsize{7}{10}\selectfont
\begin{tabular}{|l|l|l|l|l|l|}
\hline
Data & Method& M & HIT(s) & Structures & IS(\%) \\ \hline
\multirow{8}{*}{\begin{tabular}{c}  Union \\(Fig.~\ref{fig:union}) \end{tabular}}
	& Random 		&  \textbf{2465}	& 0.74			& [7]				& 0.29\\
	& Proximity 		&  2296		& 0.51			& [10]				& 0.45\\
	& LRANSAC 		&  2426 		& 0.83			& [27]				& 1.10\\
	& GMLESAC 		&  2409		& 0.17			& [29]				& 1.20\\	
	& PROSAC 		&  2400		& \textbf{0.00}	& [80]				& 3.34\\
	& Multi-GS 		&   835			& 0.27			& [51]				& 6.05  \\
	& ITKSF 		&  1546(231)		& 0.37			& [206(170)]			& 13.42(74.94)\\
	& ITKSF-S		&  1539(258)		& 0.32			& [\textbf{386}(174)]		& \textbf{24.87}(67.75) \\ \hline

\multirow{8}{*}{\begin{tabular}{c} Symon\\ (Fig.~\ref{fig:ladysymon}) \end{tabular}}
	& Random		& \textbf{2532}	& 0.87 			& [108,5]				& 4.47\\
	& Proximity		& 2363			& \textbf{0.34}	& [162,12]				& 7.34\\
	& LRANSAC		& 2490	 		& 1.09			& [141,5]				& 5.88\\
	& GMLESAC 		& 2473			& 0.77			& [264,5]				& 10.88\\
	& PROSAC 		& 2478			& 3.27			& [\textbf{722},3]			& 29.26\\
	& Multi-GS		& 997			& \textbf{0.34}	& [253,48]				& 30.22\\
	& ITKSF		& 1738(148)		& 0.37			& [513(30),125(42)]			& 36.71(52.86)\\
	& ITKSF-S		& 1737(131)		& 0.35			& [594(15),\textbf{156}(64)]		& \textbf{43.18}(52.86) \\ \hline
\multirow{8}{*}{\begin{tabular}{c} Hartley\\ (Fig.~\ref{fig:hartley}) \end{tabular}}
	& Random		& \textbf{2458}	& 4.72			& [15,0]					& 0.63\\
	& Proximity		& 2275			& 3.67			& [23,1]					& 1.06\\
	& LRANSAC		& 2411			& 4.76			& [34,0]					& 1.42\\
	& GMLESAC 		& 2395			& 3.12			& [45,1]					& 1.93\\
	& PROSAC 		& 2399			& \textbf{0.29	}	& [122,3]					& 5.23\\
	& Multi-GS 		& 846			& 1.08			& [101,5]					& 12.52\\
	& ITKSF		& 1592(286)		& 0.69			& [272(175),31(14)]				& 19.19(66.42)\\
	& ITKSF-S		& 1583(346)		& 0.57			& [\textbf{380}(181),\textbf{52}(29)]	& \textbf{27.12}(60.95)\\ \hline
\multirow{8}{*}{\begin{tabular}{c} NEEM\\ (Fig.~\ref{fig:neem}) \end{tabular}}
	& Random		& \textbf{2512}	& 2.70			& [11,2,3]							& 0.64\\
	& Proximity		& 2339			& 1.40			& [37,4,9]							& 2.13\\
	& LRANSAC		& 2470			& 2.94			& [20,3,4]							& 1.10\\
	& GMLESAC 		& 2454			& 1.21			& [22,6,5]							& 1.36\\
	& PROSAC 		& 2469			& 1.03			& [57,23,13]							& 3.76\\
	& Multi-GS 		& 1011			& 0.56			& [110,26,56]							& 18.91\\
	& ITKSF		& 1717(334)		& 0.61			& [251(128),79(16),153(115)]					& 28.14(77.71)\\
	& ITKSF-S		& 1704(334)		& \textbf{0.54}	& [\textbf{300}(116),\textbf{100}(21),\textbf{187}(131)]	& \textbf{34.42}(80.00)\\ \hline
\multirow{8}{*}{\begin{tabular}{c} Johnson\\ (Fig.~\ref{fig:johnsona}) \end{tabular}}
	& Random		& \textbf{2436}	& 3.55			& [4,9,2,1]								& 0.68\\
	& Proximity		& 2261			& 1.55			& [11,15,5,5]								& 1.61\\
	& LRANSAC		& 2386			& 3.54			& [4,18,3,2]								& 1.13\\
	& GMLESAC 		& 2373			& 3.18			& [6,12,2,2]								& 0.91\\
	& PROSAC 		& 2382			& 3.97			& [7,29,1,4]								& 1.73\\
	& Multi-GS 		& 805			& \textbf{0.50}	& [53,71,16,31]							& 21.15\\
	& ITKSF		& 1522(434)		& 0.57			& [162(84),211(129),57(7),100(83)]				& 34.83(68.71)\\
	& ITKSF-S		& 1507(385)		& \textbf{0.50}			& [\textbf{207}(76),\textbf{271}(113),\textbf{63}(7),\textbf{128}(96)]	& \textbf{44.37}(74.47)\\ \hline
 \end{tabular}
}
\end{table}

\subsection{Fundamental Matrix Estimation}
\begin{figure}[H]

\centering
\subfigure[Cube]{
\includegraphics[width=.45\linewidth]{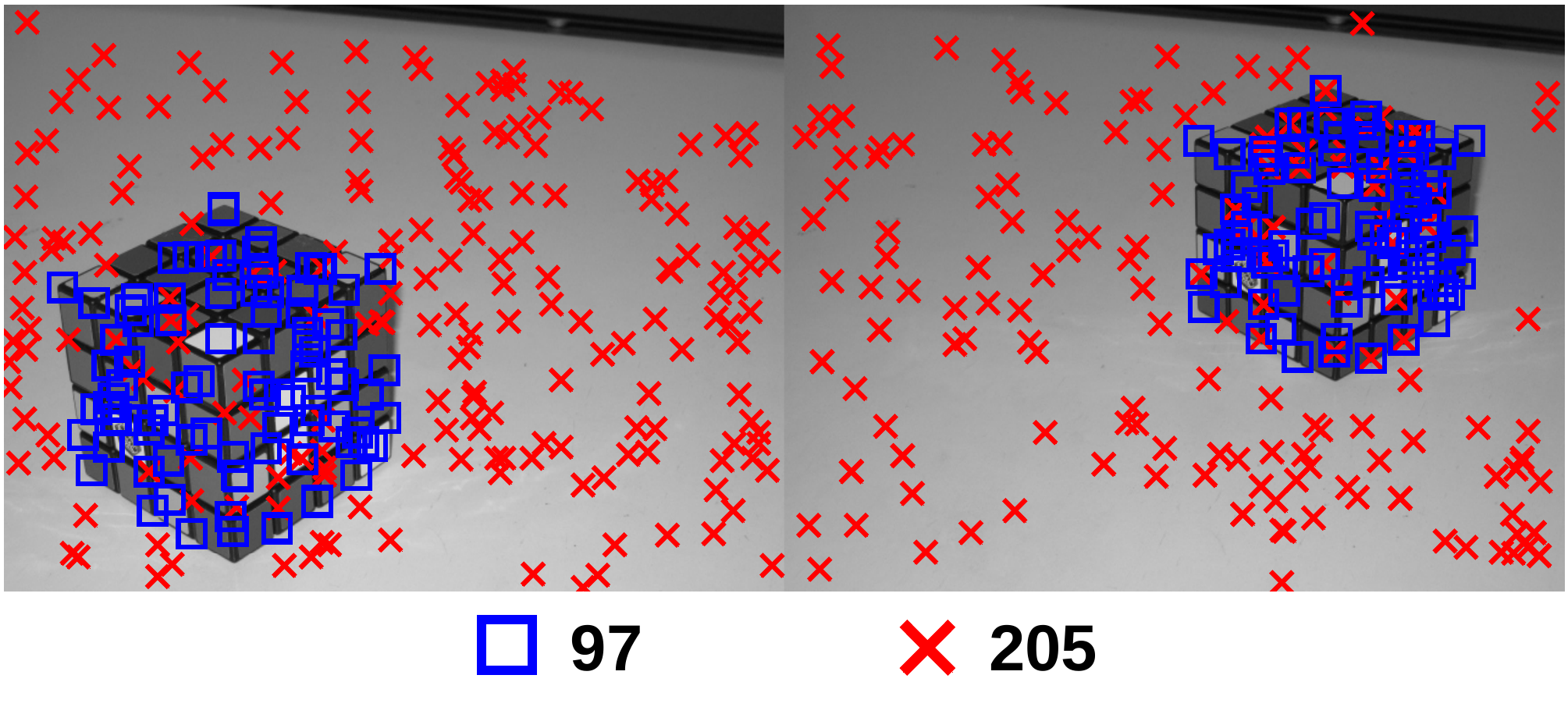}
\label{fig:cube}
}
\subfigure[Bread]{
\includegraphics[width=.45\linewidth]{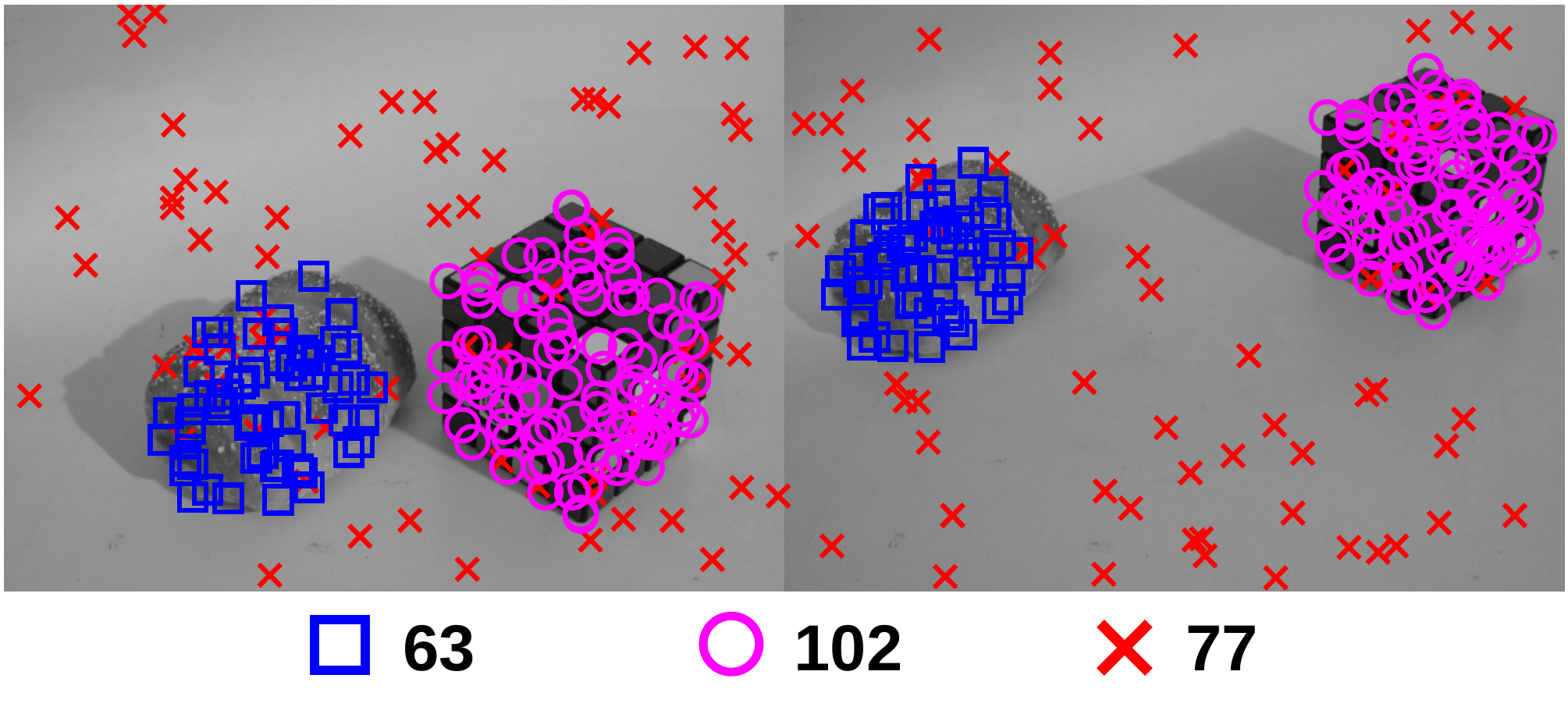}
\label{fig:breadcube}
}
\subfigure[Car]{
\includegraphics[width=.45\linewidth]{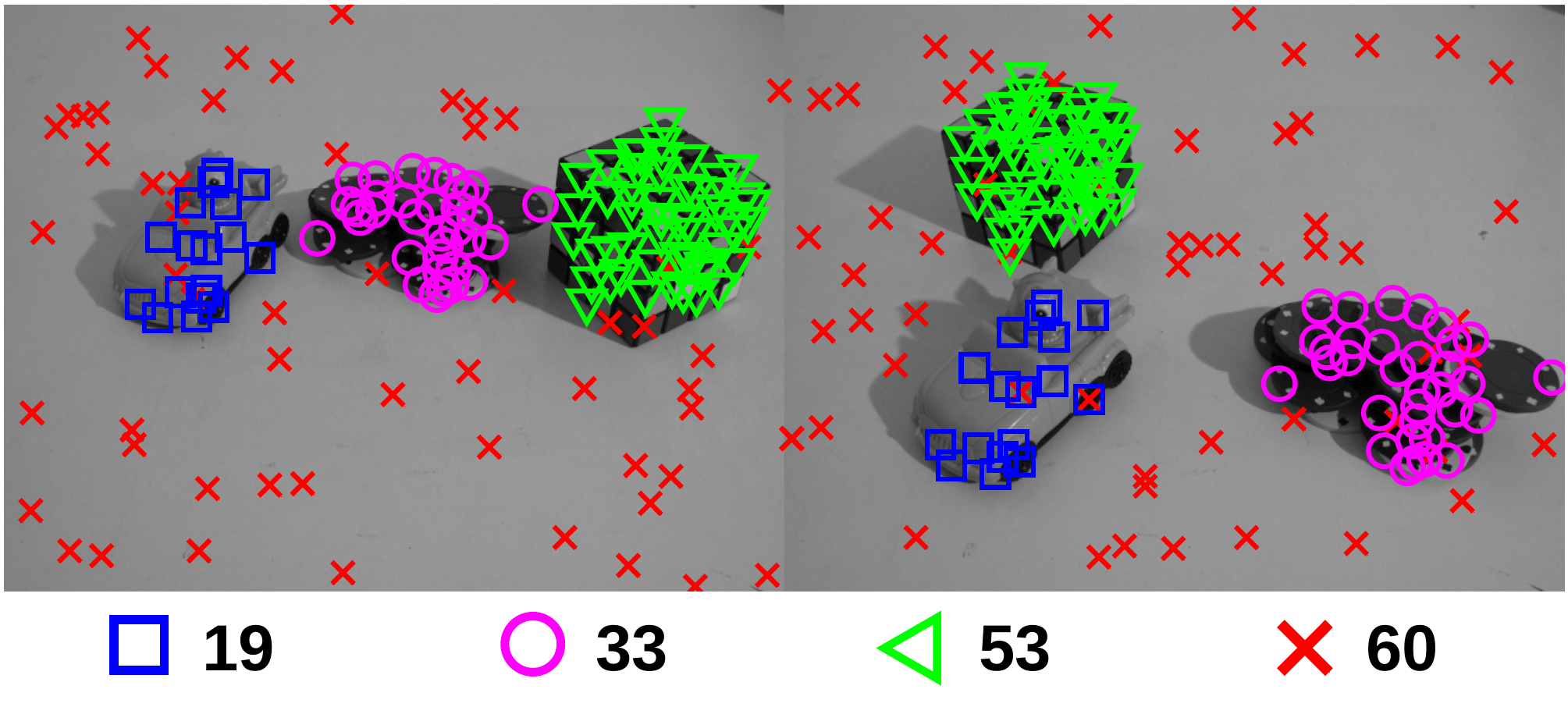}
\label{fig:carchipscube}
}
\subfigure[Chips]{
\includegraphics[width=.45\linewidth]{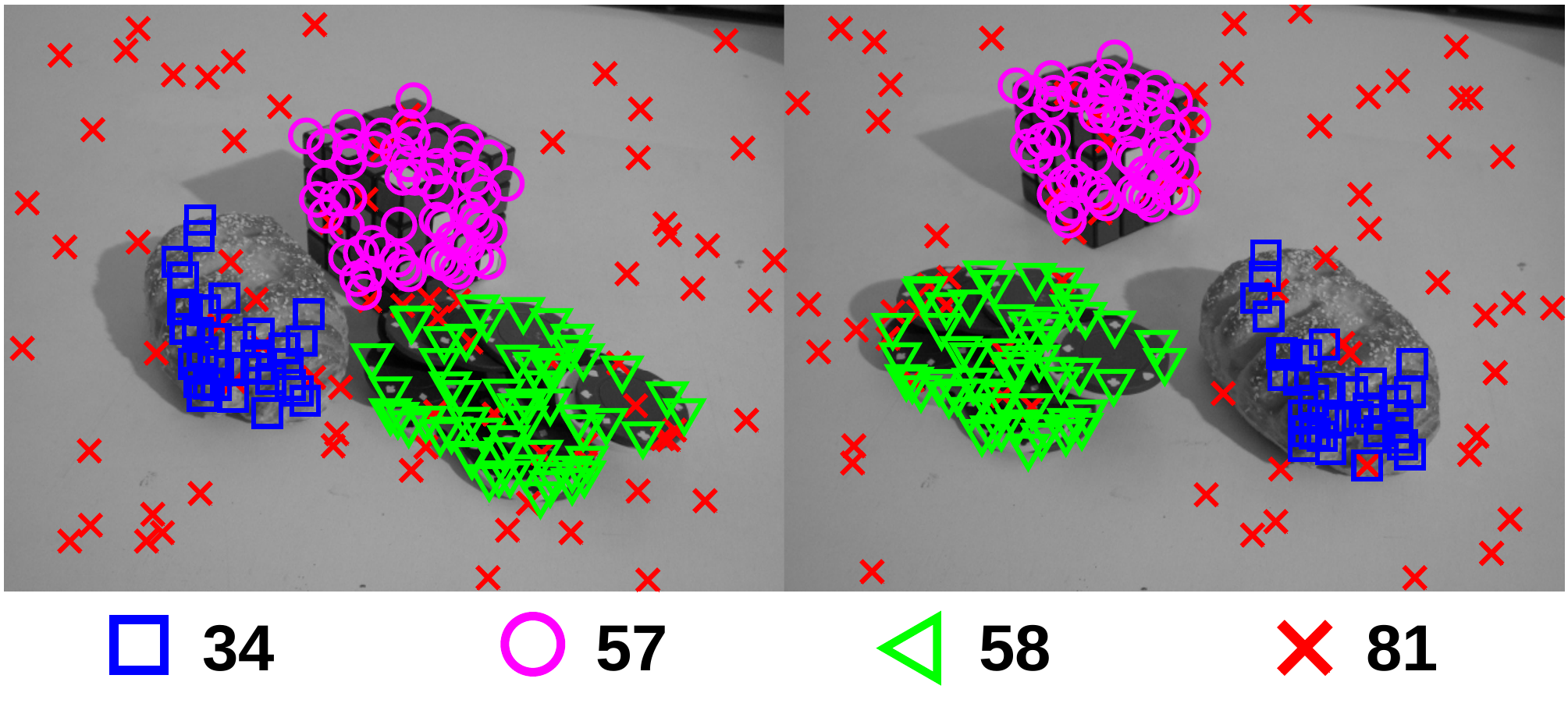}
\label{fig:breadcubechips}
}
\subfigure[ToysA]{
\includegraphics[width=.45\linewidth]{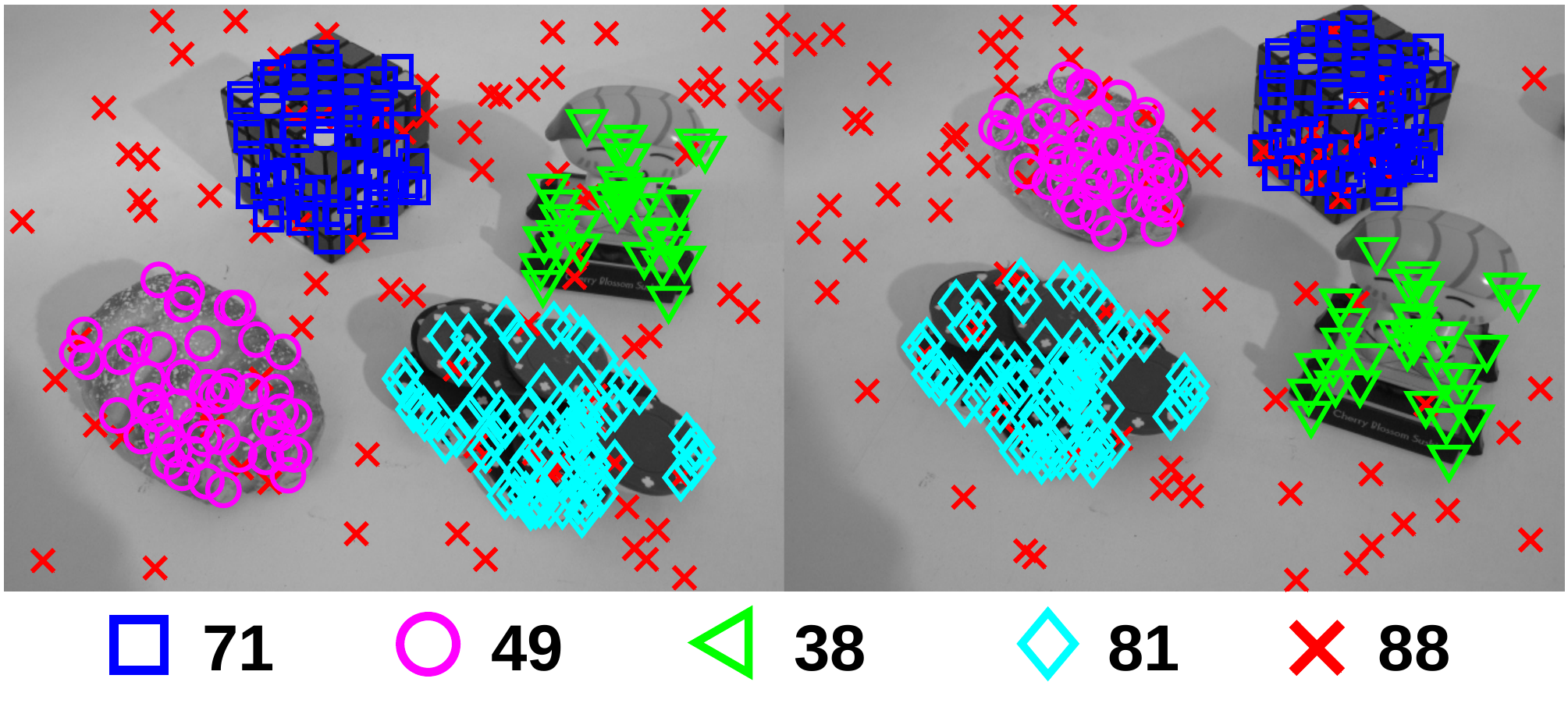}
\label{fig:cubebreadtoychips}
}
\subfigure[ToysB]{
\includegraphics[width=.45\linewidth]{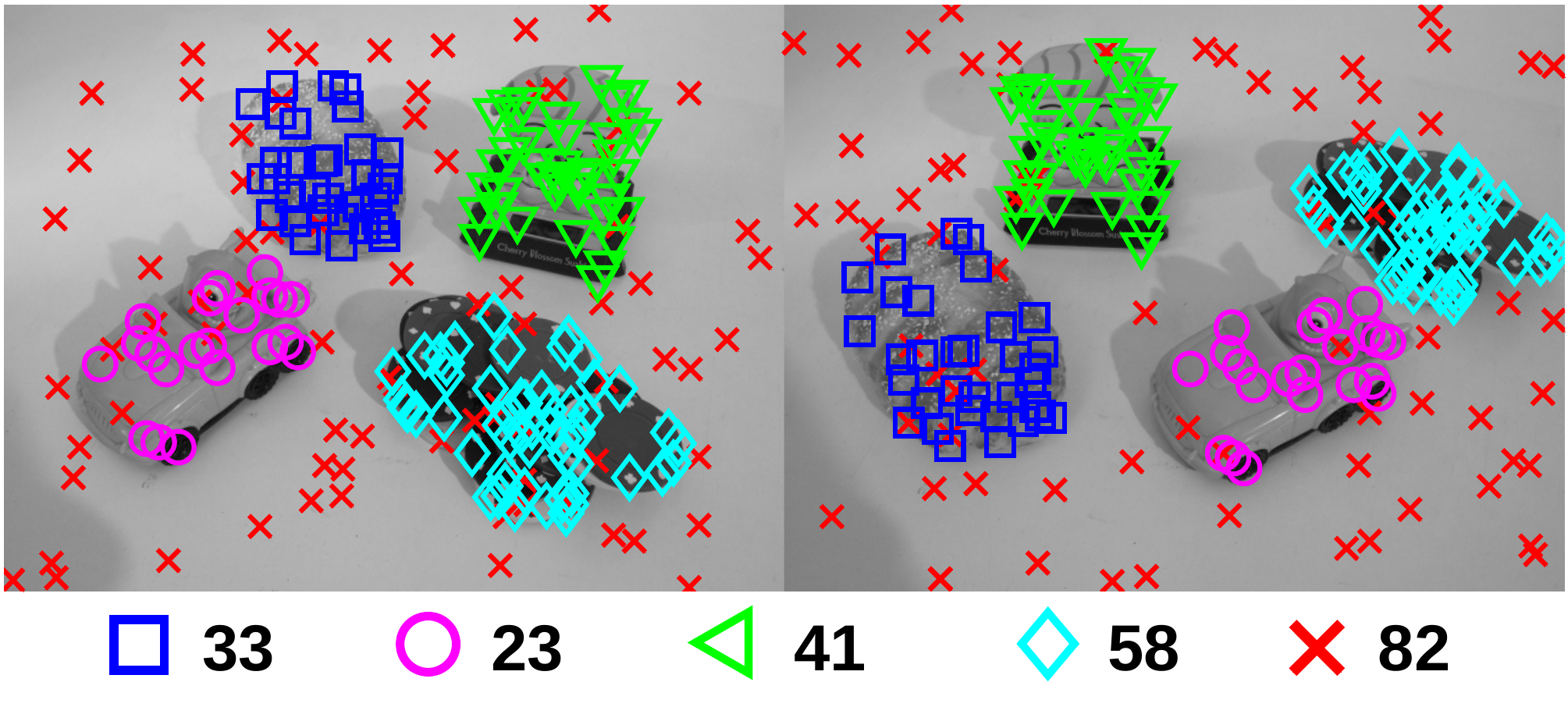}
\label{fig:breadcartoychips}
}
\caption{Data for fundamental matrix estimation. Red crosses indicate the gross outliers and other colored markers indicate the structure membership. (Best viewed in color)} 
\label{fig:fundadata}
\end{figure}

We also evaluate the proposed method on fundamental matrix estimation. The data used for this experiment is shown in Fig.~\ref{fig:fundadata}. We use the 8-point algorithm \cite{multiview04} to estimate a fundamental matrix\footnote{{\small\url{http://www.robots.ox.ac.uk/~vgg/hzbook/code/}}}. Each method is given 50 random runs and each for 10 CPU seconds.

Table~\ref{tab:fundatable} shows the performance of guided sampling methods and also the hypothesis filtering result from our proposed method.

\begin{table}[H]
  \centering
\caption{Performance of guided sampling methods in fundamental matrix estimation. The notation is same as Table~\ref{tab:homotable}.}
\label{tab:fundatable}
{\fontsize{6}{10}\selectfont
\begin{tabular}{|l|l|l|l|l|l|}
\hline
Data & Method& M & HIT(s)  & Structures & IS(\%) \\ \hline
\multirow{8}{*}{\begin{tabular}{c} Cube\\ (Fig.~\ref{fig:cube}) \end{tabular}}
	& Random		& \textbf{5083}	& 8.19			& [1]				& 0.01\\
	& Proximity		& 4741			& 3.29			& [3]				& 0.06\\
	& LRANSAC		& 4950			& 8.28			& [1]				& 0.01\\
	& GMLESAC 		& 4960			& 5.08			& [2]				& 0.03\\
	& PROSAC 		& 4904			& 1.73			& [3]				& 0.05\\
	& Multi-GS 		& 981			& \textbf{0.30}	& [134]			& 13.68\\
	& ITKSF		& 2903(654)		& 0.34			& [483(338)]			& 16.63(51.98)\\
	& ITKSF-S		& 2896(636)		& 0.37			& [\textbf{532}(334)]		& \textbf{18.36}(52.74)\\ \hline

\multirow{8}{*}{\begin{tabular}{c} Bread\\ (Fig.~\ref{fig:breadcube}) \end{tabular}}
	& Random		& \textbf{6130}	& 9.41			& [0,6]						& 0.09	\\
	& Proximity		& 5640			& 3.06			& [4,52]					& 1.00\\
	& LRANSAC		& 5926			& 9.74			& [0,8]						& 0.13\\
	& GMLESAC 		& 5973			& 8.66			& [0,32]					& 0.54\\
	& PROSAC 		& 5855 		& 8.46			& [0,28]					& 0.48\\
	& Multi-GS 		& 1108			& \textbf{0.24}	& [129,244]					& 33.69\\
	& ITKSF		& 3402(545)		& 0.32			& [506(220),867(144)]				& 40.36(69.48)\\
	& ITKSF-S		& 3393(497)		& 0.31			& [\textbf{534}(188),\textbf{933}(153)]	& \textbf{43.34}(66.88)\\ \hline

\multirow{8}{*}{\begin{tabular}{c} Car\\ (Fig.~\ref{fig:carchipscube}) \end{tabular}}
	& Random		& \textbf{8277}	& 10.00		& [0,0,1]							& 0.01\\
	& Proximity		& 7449			& 10.00		& [0,0,11]							& 0.16\\
	& LRANSAC		& 7903			& 10.00		& [0,0,1]							& 0.01\\
	& GMLESAC 		& 8026			& 10.00		& [0,0,11]							& 0.14\\
	& PROSAC 		& 7782			& 10.00		& [0,0,6]							& 0.08\\
	& Multi-GS 		& 1382			& 0.92			& [18,81,282]							& 27.56\\
	& ITKSF		& 4212(520)		& \textbf{0.77}	& [\textbf{157}(47),319(12),996(361)]					& 34.95(81.08)\\
	& ITKSF-S		& 4198(531)		& 0.91			& [96(30),\textbf{374}(25),\textbf{1270}(369)]	& \textbf{41.45}(80.16)\\ \hline

\multirow{8}{*}{\begin{tabular}{c} Chips\\ (Fig.~\ref{fig:breadcubechips}) \end{tabular}}
	& Random		& \textbf{6361}	& 10.00		& [0,0,0]							& 0.00\\
	& Proximity		& 5836			& 10.00		& [0,1,0]							& 0.02\\
	& LRANSAC		& 6141			& 10.00		& [0,0,0]							& 0.01\\
	& GMLESAC 		& 6180			& 10.00		& [0,1,0]							& 0.02\\
	& PROSAC 		& 6065			& 10.00		& [0,1,0]							& 0.02\\
	& Multi-GS 		& 1142			& 0.64			& [37,135,127]						& 26.17\\
	& ITKSF		& 3405(812)		& \textbf{0.49}	& [184(83),521(263),504(187)]				& 35.53(65.55)\\
	& ITKSF-S		& 3395(736)		& \textbf{0.49}		& [\textbf{190}(81),\textbf{534}(238),\textbf{546}(157)]	& \textbf{37.42}(64.82)\\ \hline

\multirow{8}{*}{\begin{tabular}{c} ToysA\\ (Fig.~\ref{fig:cubebreadtoychips}) \end{tabular}}
	& Random		& \textbf{4723}	& 10.00		& [0,0,0,0]								& 0.00\\
	& Proximity		& 4415			& 10.00		& [0,0,0,1]								& 0.03\\
	& LRANSAC		& 4606			& 10.00		& [0,0,0,0]								& 0.00\\
	& GMLESAC 		& 4629			& 10.00		& [0,0,0,0]								& 0.00\\
	& PROSAC 		& 4562			& 10.00		& [0,0,0,1]								& 0.03\\
	& Multi-GS 		& 932			& 1.04			& [87,36,23,110]							& 27.48\\
	& ITKSF		& 2793(627)		& \textbf{0.71}	& [393(209),\textbf{232}(123),154(104),451(18)]			& 44.05(72.48)\\
	& ITKSF-S		& 2762(544)		& 0.79			& [\textbf{437}(171),213(104),\textbf{160}(108),\textbf{445}(17)]	& \textbf{45.40}(73.42)\\ \hline

\multirow{8}{*}{\begin{tabular}{c} ToysB\\ (Fig.~\ref{fig:breadcartoychips}) \end{tabular}}
	& Random		& \textbf{6202}	& 10.00		& [0,0,0,0]								& 0.00\\
	& Proximity		& 5706			& 10.00		& [0,0,0,1]								& 0.02\\
	& LRANSAC		& 5990			& 10.00		& [0,0,0,0]								& 0.00\\
	& GMLESAC 		& 6040			& 10.00		& [0,0,0,0]								& 0.00\\
	& PROSAC 		& 5926			& 10.00		& [0,0,0,0]								& 0.00\\
	& Multi-GS 		& 1104			& 2.16			& [39,8,82,104]							& 21.03\\
	& ITKSF		& 3406(637)		& \textbf{1.13}	& [214(109),\textbf{95}(40),361(271),448(34)]			& 32.84(71.78)\\
	& ITKSF-S		& 3399(664)		& 1.56			& [\textbf{238}(132),72(32),\textbf{403}(273),\textbf{485}(29)]	& \textbf{35.25}(70.06)\\ \hline

 \end{tabular}
}
\end{table}

\section{Conclusions}
\label{sec:conclusion}
We propose a novel guided sampling scheme based on the distances between top-$k$ lists that are derived from residual sorting information. In 
contrast to many existing sampling enhancement techniques, our method does not rely on any domain-specific knowledge, and is capable of handling 
multiple structures. Moreover, while performing sampling, our method simultaneously filters the hypotheses such that only a small but very promising 
subset remains. This permits the use of simple agglomerative clustering on the surviving hypotheses for accurate model selection. Experiments on 
synthetic and real data show the superior performance of our approach over previous methods.

\bibliographystyle{elsarticle-num}
\bibliography{itksf-journal}











\end{document}